%% file: Main.tex
\documentclass[preprint,12pt]{elsarticle}

\usepackage[utf8]{inputenc}
\usepackage{graphicx}
\usepackage{framed}
\usepackage{amssymb}
\usepackage{lineno}
\usepackage{xcolor}
\usepackage{caption}
\usepackage{doi}
\usepackage[title,titletoc]{appendix}
\usepackage{booktabs}
\usepackage{setspace}
\usepackage{bigstrut}
\usepackage{afterpage}
\usepackage{array}
\usepackage{multirow}
\usepackage{amsfonts,epsfig}
\usepackage{rotating}
\usepackage{algorithmicx}
\usepackage{color}
\usepackage{subcaption}
\usepackage[normalem]{ulem}
\usepackage{flushend}
\usepackage{lscape}
\usepackage[noend]{algpseudocode}
\usepackage{epstopdf}
\usepackage{amsmath,mathtools}
\usepackage{framed}
\usepackage{algorithm}
\usepackage{hyperref}
\usepackage[margin=0.7 in]{geometry}
\usepackage{float}
\usepackage[symbol]{footmisc}
\usepackage{comment}
\usepackage{natbib}

\algnewcommand\algorithmicswitch{\textbf{switch}}
\algnewcommand\algorithmiccase{\textbf{case}}
\algdef{SE}[SWITCH]{Switch}{EndSwitch}[1]{\algorithmicswitch\ #1\ \algorithmicdo}{\algorithmicend\ \algorithmicswitch}%
\algdef{SE}[CASE]{Case}{EndCase}[1]{\algorithmiccase\ #1}{\algorithmicend\ \algorithmiccase}%
\algtext*{EndSwitch}%
\algtext*{EndCase}%

\journal{Transportation Research Part C: Emerging Technologies}

\begin{document}
\sloppy
\begin{frontmatter}

\title{\Large \textbf{A new Hyper-heuristic based on Adaptive Simulated Annealing and Reinforcement Learning for the Capacitated Electric Vehicle Routing Problem}}

\author[a]{Erick Rodr\'iguez-Esparza*}
\ead{erick.rodriguez@deusto.es (corresponding author)}
\author[a,b]{Antonio D Masegosa}
\author[c]{Diego Oliva}
\author[a]{Enrique Onieva}

\address[a]{DeustoTech, Faculty of Engineering, University of Deusto, Av. Universidades, 24, 48007 Bilbao, Spain.}
\address[b]{Ikerbasque, Basque Foundation for Science, Plaza Euskadi, 5, 48009 Bilbao, Spain.}
\address[c]{Depto. de Innovaci\'on Basada en la Informaci\'on y el Conocimiento, Universidad de Guadalajara, CUCEI, Av. Revoluci\'on 1500, C.P. 44430, Guadalajara, Jal.}

\begin{abstract}
    Electric vehicles (EVs) have been adopted in urban areas to reduce environmental pollution and global warming as a result of the increasing number of freight vehicles. However, there are still deficiencies in routing the trajectories of last-mile logistics that continue to impact social and economic sustainability. For that reason, in this paper, a hyper-heuristic (HH) approach called Hyper-heuristic Adaptive Simulated Annealing with Reinforcement Learning (HHASA$_{RL}$) is proposed. It is composed of a multi-armed bandit method and the self-adaptive Simulated Annealing (SA) metaheuristic algorithm for solving the problem called Capacitated Electric Vehicle Routing Problem (CEVRP). Due to the limited number of charging stations and the travel range of EVs, the EVs must require battery recharging moments in advance and reduce travel times and costs. The HH implemented improves multiple minimum best-known solutions and obtains the best mean values for some high-dimensional instances for the proposed benchmark for the IEEE WCCI2020 competition.

\end{abstract}

\begin{keyword}

Last-mile logistics \sep Hyper-heuristic \sep Electric vehicles \sep Capacitated Electric Vehicle Routing Problem \sep Combinatorial optimization \sep Reinforcement Learning

\end{keyword}

\end{frontmatter}

\input{1Introduction}
\input{2Problem}
\input{3Related}
\input{4Background}
\input{5Algorithm}
\input{6Experimental}
\input{7Results}
\input{8Conclusions}

\section*{CRediT authorship contribution statement}
\textbf{Erick Rodr\'iguez-Esparza:} Conceptualization, Methodology, Software, Writing – original draft. \textbf{Antonio D Masegosa:} Formal analysis, Supervision,
Writing – review \& editing. \textbf{Diego Oliva:} Writing – review \& editing.
\textbf{Enrique Onieva:} Visualization, Supervision, Writing – review \& editing.

\section*{Acknowledgements}
This research has been supported by the University of Deusto Research Training Grants Programme and by the Spanish Ministry of Science and Innovation through research project PID2019-109393RA-I00.

This research has also been supported by European Union’s Horizon 2020 research and innovation programme under grant agreement No. 861540 [project SENATOR (Smart Network Operator Platform enabling Shared, Integrated and more Sustainable Urban Freight Logistics)].

\bibliographystyle{model2-names}
\biboptions{authoryear}
\setlength{\bibsep}{2pt plus 0.3ex}
\bibliography{Biblio.bib}

\end{document}

%% file: 1Introduction.tex
\section{Introduction}
\label{S:1}

Over the last recent years, there has been a remarkable increase in the use of e-commerce systems around the world, which in turn has had an impact on distribution and last-mile strategies \citep{viu2020impact, ignat2020commerce, castillo2018crowdsourcing}. In 2018, a growth rate of 23.3\% was reported worldwide \citep{patella2021adoption}, and these numbers have increased dramatically as a result of the COVID-19 pandemic situation, which has led to huge volumes of packages being delivered daily. Some research reported that some business websites perceived a 74\% increase in the e-commerce rate, while 52\% of customers avoided in-store purchases \citep{giuffrida2022optimization, singh2021impact, bhatti2020commerce}.

The last-mile is the last journey of a good to be delivered to the end customer. It has been given a greater focus on logistics strategies because it strongly influences customer satisfaction, time, delivery cost, and ease of use \citep{archetti2021recent, vakulenko2019service}. Due to the increasing number of freight vehicles in urban areas, last-mile logistics operations have a considerable impact on three different aspects of sustainability: economic (efficiency and delivery costs), social (congestion and health problems) and environmental (CO$_2$ emissions and noise pollution) \citep{janjevic2019integrated}. One of the measures that have been taken to address the environmental impact is the adoption of electric vehicles (EVs), since the transportation sector is estimated to be responsible for about 20-25\% of global C0$_2$ emissions \citep{su13010226, bosona2020urban, yi2018energy, he2018optimal}.

Furthermore, in line with this need to improve the efficiency of last-mile logistics, applying the Vehicle Route Problem (VRP) concept to this field began to gain importance in the research community. Its main objective is to optimize the routes for the transport of goods from one or several warehouses to a set of geographically dispersed clients to increase efficiency and reduce time and/or costs; in this way, it is intended to combat the social and economic impact \citep{zirour2008vehicle}. However, the classical models do not provide an answer to the particularities of increasing the use of EVs in last-mile logistics that we mention above. The main reason is that they do not consider restrictions in terms of vehicle autonomy and the need to recharge the batteries. From this, the Electric Vehicle Problem (EVRP) concept arises, which is a variation of the conventional VRP.

The approaches proposed for solving VRP, EVRP, and variants can be classified as: exact algorithms, heuristics, metaheuristics, and HHs \citep{asghari2020green,blocho2020heuristics}. 
An exact algorithm always finds the optimal solution, but it has the disadvantage that it can only tackle problems of relatively small size, due to its high computational time requirements \citep{purkayastha2020study}. Metaheuristic algorithms are comprehensive techniques that provide a general structure and strategic criteria for developing a heuristic method to solve the problem \citep{oliva2020balancing,fausto2020ants,morales2020better}. Heuristics and metaheuristics are good alternatives for solving VRP and its variations, as the previously mentioned EVRP. Nevertheless, it is challenging to decide when to apply a specific method or operator \citep{osaba2020vehicle}. Moreover, there are still difficulties in using the current algorithms due to the existence of multiple parameters and the high sensitivity in their configuration \citep{swiercz2017hyper}. Therefore, the choice of a proper search mechanism is crucial since it allows obtaining optimal results with high accuracy. In this sense, the use of hyper-heuristics (HH) is becoming popular. HH are considered high-level automatic search methods that work combining, generating and selecting low-level heuristics to solve computationally complex problems \citep{burke2013hyper}. 

HH algorithms generally involve two main components, named selection and move acceptance. The selection mechanism helps to identify which element from the pool of low-level heuristics that should be applied at each stage of the solving process \citep{scoczynski2021selection}. The move acceptance mechanism decides whether to accept or discard the generated solution \citep{turky2018hyper}. HHs are powerful tools, however, the design of heuristic selection is complex and time-consuming; because it is usually done by trial and error that depends on the problem \citep{zhang2021deep,drake2020recent,choong2018automatic}.

To overcome such problems, the use of machine learning mechanisms, including Reinforcement Learning (RL) strategies could be seen as a good alternative to create smart and self-adaptive algorithms \citep{wang2021deep,de2020hyper,largo2020green}. RL methods are learning strategies to find the best actions to apply for a particular state or observation of/from an environment. The agent chooses actions sequentially in discrete time steps from a set of available actions and receives a reward that varies depending on the utility of the action taken. The actions taken affects or change the state of the environment in which the agent operates.

Based on the ideas outlined above, in this paper, we propose a HH to address a variant of the EVRP, called the Capacitated Electric Vehicle Routing Problem (CEVRP) efficiently. The main components of the proposed algorithm, called Hyper-heuristic Adaptive Simulated Annealing and Reinforcement Learning (HHASA$_{RL}$), are a RL algorithm as a selection mechanism and the Metropolis criterion of the well-known Simulated Annealing (SA) metaheuristic algorithm as the movement acceptance mechanism. The problem of selection among the pool of heuristics is treated as a multi-armed bandit problem \citep{slivkins2019introduction}. This stochastic problem is used for the dilemma of exploration and exploitation of the algorithm. It is assigned a fixed set of options or actions, and the agent selects one within that set to maximize the long-term cumulative reward.

A comparison of three of the most commonly used algorithms for dealing with multi-armed bandit problem is presented in this paper. The RL algorithms used for this comparison are Epsilon Greedy ($\epsilon$-$G$), Thompson Sampling ($TS$) and Upper Confidence Bound 1 ($UCB1$) to identify which of them most efficiently selects the low-level heuristic for the CEVRP. These techniques guide and control the local search of the SA by choosing the heuristic that is best applied during the iterative process of the algorithm to optimize the long-term performance by improving the quality of the solutions.

The most relevant contributions of this proposal can be summarized as follows:
\begin{itemize}
    \item A methodology for solving high-dimensional electric vehicle routing problems.
    \item A hyper-heuristic based on hybridization of self-adaptive simulated annealing and reinforcement learning treating it as a multi-armed bandit problem. 
    \item State-of-the-art results for the CEVRP benchmark proposed for the IEEE WCCI2020 competition.
\end{itemize}

For the validation of the performance and competitiveness of this proposal, we have used instances of the CEVRP from the IEEE WCCI 2020 competition. This benchmark contains 17 instances of short and long problems, ranging from 21 to 1000 customers. Statistical analysis and non-parametric test were conducted to validate results of the proposed algorithm in comparison with the rest of the approaches.

The remainder of this article is organized as follows: Section \ref{S:2} provides a formal description of the CEVRP.
Then, Section \ref{S:3} shows the related works to solve CEVRP. Section \ref{S:4} gives background information about important concepts used for the proposed HH. Section \ref{S:5} details the proposed HHASA$_{RL}$ algorithm. Afterward, Section \ref{S:6} presents the experimental framework. In Section \ref{S:7}, the experiments and results are shown. Finally, in Section \ref{S:8} the conclusions and further work are included.

%% file: 2Problem.tex
\section{The Capacitated Electric Vehicle Routing Problem}
\label{S:2}

The CEVRP is a $\mathcal{NP}$-hard combinatorial optimization problem that is a variation of the conventional VRP, where capacity constraints are applied. In the problem, given a fleet of EVs with a specific load capacity and battery level capacity, the objective is to find optimal routes to fulfill the demand of a set of customers that minimize the total traveling distance of each vehicle, subject to different constraints. The main difference between VRP and EVRP is that EVRP considers the specific characteristics of EVs such as the autonomy of the vehicle and the need to recharge the batteries. \citep{mavrovouniotis2018ant, erdelic2019survey}.

The CEVRP is defined on a complete, undirected graph $G(V,A)$, where $V=\{ D\cup C \cup S \}$ and $A=\{ (i,j) \mid i,j \in V, i \neq j \}$. $V$ denotes the set of nodes composed of a set $C$ of $n_c$ customers, a set $S$ of $n_s$ external charging stations, and a central depot denoted by $D$. The set of arcs connecting the mentioned nodes is denoted by $A$. Each arc is associated with a non-negative value $d_{ij}$ that represents the distance between nodes $i$ and $j$. When an arc $(i,j)$ is travelled by an electric vehicle, it consumes an amount of energy $e_{i,j}=h\cdot d_{ij}$, where the parameter $h$ denotes the energy consumption rate.

Furthermore, each customer $i$ has a specific delivery demand $q_{i}$. All electric vehicles are identical, and each has a maximum capacity of load demand ($Max_C$) and a maximum battery capacity ($Max_Q$), and these parameters should not be exceeded for each EV. In addition, they start (fully loaded and charged) and end at the depot; it is important to mention that each of the vehicles can visit the charging stations several times, but all customers must be visited exactly once.

The CEVRP mathematical model is formulated as follows \citep{mavrovouniotis2020benchmark2}:

\begin{subequations} 
\begin{align}
&  & min  \quad f(\mathbf{x}) &= \sum_{i \in V, j \in V, i \neq j}d_{ij} \cdot x_{ij}, & \label{Eq:1}\\[0.1cm]
& s.t. & \sum_{j \in V, i \neq j} x_{ij} &= 1, & \forall i \in C, \label{Eq:2}\\[0.1cm]
&  & \sum_{j \in V, i \neq j} x_{ij} &\leq 1, & \forall i \in S, \label{Eq:3}\\[0.1cm]
&  & \sum_{j \in V, i \neq j} x_{ij} &- \sum_{j \in V, i \neq j} x_{ji} = 0, & \forall i \in V, \label{Eq:4}\\[0.1cm]
&  & u_j \leq u_i &- c_{i} \cdot x_{ij} + Max_C \cdot (1-x_{ij}), & \forall i \in V, \forall j \in V, i \neq j, \label{Eq:5}\\[0.1cm]
& & 0 \leq u_i &\leq Max_C, & \forall i \in V, \label{Eq:6}\\[0.05cm]
& & y_j \leq y_i &- hd_{ij}\cdot x_{ij} + Max_Q \cdot (1-x_{ij}), & \forall i \in I, \forall j \in V, i \neq j, \label{Eq:7}\\[0.1cm]
& & y_j \leq Max_Q &- hd_{ij} \cdot x_{ij}, \forall i \in S \cup \left\lbrace 0 \right\rbrace, & \forall j \in V, i \neq j, \label{Eq:8}\\[0.1cm]
& & 0 \leq y_i &\leq Max_Q, & \forall i \in V, \label{Eq:9}\\[0.1cm]
& & x_{ij} &\in \left\lbrace 0,1 \right\rbrace, & \forall i \in V, \forall j \in V, i \neq j, \label{Eq:10}
\end{align}
\label{Eq:a}
\end{subequations}

where Eq. (\ref{Eq:1}) defines the CEVRP objective function, which is to minimize the total travel distance of all EVs. The restriction of Eq. (\ref{Eq:2}) refers to the fact that each customer must be served only once. On the other hand, Eq. (\ref{Eq:3}) indicates that the charging stations can be visited several times. Eq. (\ref{Eq:4}) establishes flow conservation by guaranteeing that at each node, the number of incoming arcs is equal to the number of outgoing arcs. The Eqs. (\ref{Eq:5}) and (\ref{Eq:6}) are the capacity constraints that guarantee that the load of an EV is non-negative upon arrival at any node, including the deposit. And the energy constraints of Eqs. (\ref{Eq:7}), (\ref{Eq:8}) and (\ref{Eq:9}) ensure that the battery charge level never drops below 0. Finally, Eq. (\ref{Eq:10}) defines the set of binary decision variables ($x_{ij}$), if the arc $(i,j)$ is traveled by an EV, $x_{ij}$ is equal to one and zero otherwise.

The variables $u_i$ and $y_i$, respectively, represent the remaining charge capacity and the remaining energy level of the EV when it reaches the node $ i \in V $. Despite the explicit restrictions shown by the formulas, they also imply that all EVs must depart and return from the depot.

\begin{figure}[http]
\centering
\includegraphics[scale=0.15]{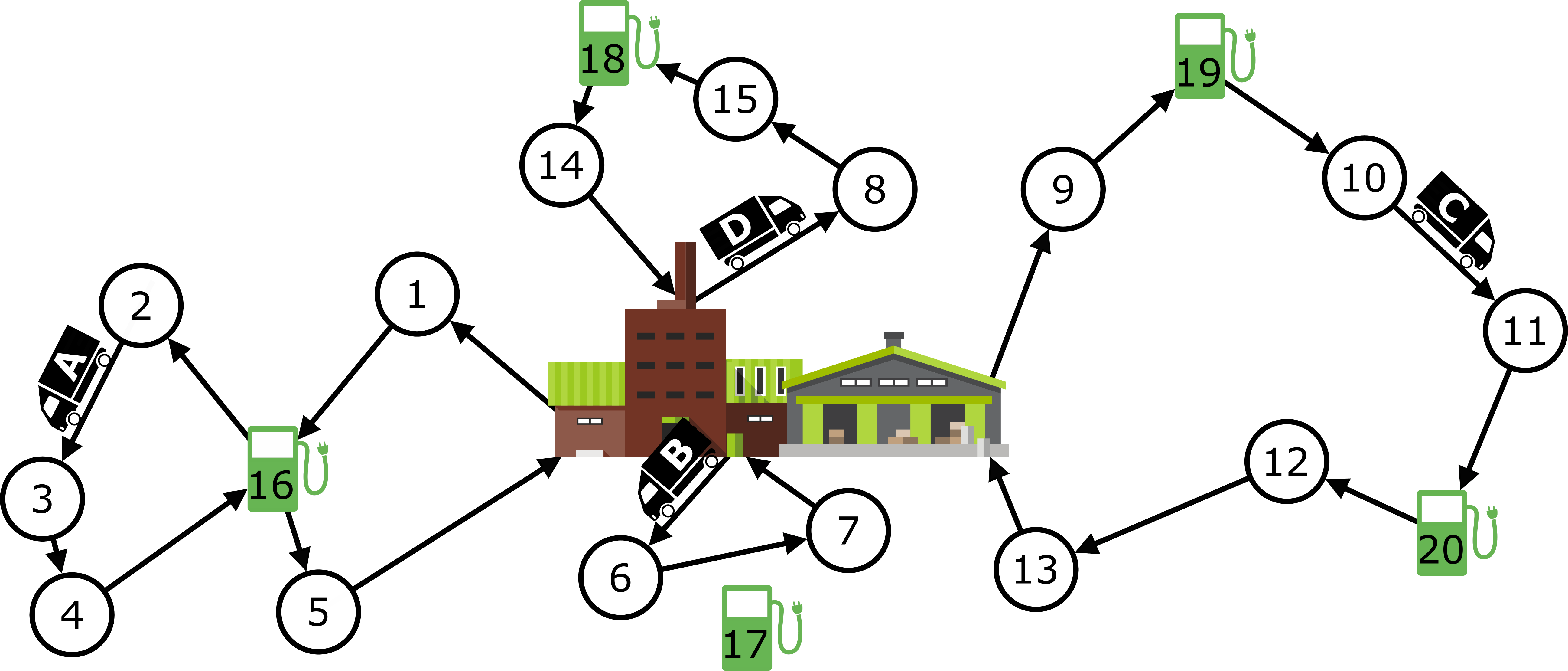}
\caption{Example of CEVRP with four routes. (A) stops twice in the same charging station, (B) does not stop in any station, (C) stops in two different stations and (D) stops in a single station.}
\label{fig:evrp}
\end{figure}

Figure \ref{fig:evrp} shows an example of four EVs (denoted by $A$, $B$, $C$, $D$) in a CEVRP problem. The figure shows different cases that can occur in the routes. In the first case, we can observe that route $A$ visits the same charging station twice. However, the opposite case would be that the EV will not run out of power, so it would not need to go through any station, as exemplified in route $B$. In route $C$, we can see that in the same way, the EV makes a long trip and needs to pass through two different stations during the journey. Finally, the EV only passes through the charging station once in route $D$ and returns to the depot.

%% file: 3Related.tex
 \section{Related Work}
\label{S:3}

In this section, some significant works proposed in the state-of-the-art to solve CEVRP are described. This problem can be seen as a variant of the green vehicle routing problem first proposed in 2012 by Erdogan and Miller-Hooks \citep{erdougan2012green}. The authors used a modified Clarke and Wright savings heuristic and a density-based clustering algorithm to find initial solutions, followed by an optimization phase. The green vehicle routing problem generally refers to routing alternative fuel vehicles, including EVs, which present the characteristic of having a limited travel range. Therefore, they require planning for battery recharging during the delivery route.

Over recent years, the scientific literature on EVRP has grown uninterrupted. In 2015, Pelletier et al. presented a detailed study of the variants of the types of EVRP \citep{pelletier201650th}. Later in 2019, Erdelić and Carić conducted a survey in which they classified articles according to the composition of the fleet, the terms of the objective function, the presence of multiple recharging technologies, and limitations such as capacities of the EVs and time windows for the customers. In addition, their work reviewed the exact, heuristic, metaheuristic and hybrid approaches applied to solve different variants of EVRP \citep{erdelic2019survey}. As can be seen from the works mentioned above on the state-of-the-art, there are different strategies to solve problems with variations in EVs. As is the case in the work of Scheinder et al., in which they extended the model to EVRP by adding time windows for the customers and solved it through a hybridization between the VNS algorithm with a TS heuristic \citep{schneider2014electric}. Their proposal was tested in 56 instances based in the benchmark proposed by Solomon \citep{solomon1987algorithms} with up to 100 clients.

Keskin and Çatay used the same instances of the Solomon benchmark \citep{keskin2016partial}. They considered the cases with a partial recharge of the battery when stopping at a recharging station and different objectives. They proposed two works using an ALNS approach by introducing new route modification heuristics to add/remove clients and stations \citep{keskin2018matheuristic}. Montoya et al. proposed an automated repair strategy that inserts stations into routes to ensure route viability and a modified multi-space sampling heuristic, which was tested on 52 cases with up to 500 clients \citep{montoya2016multi}. One year later, Montoya et al. formulated a recharging of battery process as a nonlinear function and presented a hybrid metaheuristic combining an iterated local search and a heuristic concentration to solve the new variation of the problem called EVRPNL \citep{montoya2017electric}. This proposal was tested on 120 problems, where customers ranged from 10 to 320. Recently, Mao et al. investigated a new variation of EVRP, named EVRPTW\&MC, in which they added decisions on multiple recharging options, which are partial recharging and battery swapping \citep{mao2020electric}. An improved ACO algorithm was proposed that is combined with an insertion heuristic and an improved local search. The ACO was compared and validated using 56 experimental instances used in others works in the literature  \citep{schneider2014electric}. 

On the other hand, there are also some proposals in the state-of-the-art where the researchers use RL techniques to solve the EVRP and its variations due to the advantages mentioned in Section \ref{S:1}. As is the case of Shi et al., that introduced a new off-policy RL framework with decentralized learning and centralized decision making processes used to solve the EVRP in the ride-hailing services \citep{shi2019operating}. In 2020, Lin et al. proposed a deep RL framework for solving the EVRPTW that used an attention model with a pointer network and a graph embedding layer to parameterize a stochastic policy. The REINFORCE gradient estimator is used to train the model, which has a greedy roll-out baseline. Their proposal was put to the test in seven small problems with client numbers ranging from 5 to 100 \citep{lin2020deep}. Also, Zhao et al. implemented a hybridization of a deep RL model composed of an actor, an adaptive critic, and a routing simulator with a local search strategy to increase solution quality even more. This proposal was tested on three datasets with up to 100 clients to solve the VRP and VRP with time windows \citep{zhao2020hybrid}. Recently, Bogyrbayeva et al. have developed an RL framework-based method to free-floating electric vehicle sharing systems, in which a central controller determines the routing strategies of a fleet of various shuttles. They train a recurrent neural network using a policy gradient method and compare the outcomes to heuristic solutions. For this investigation, they employed three problems with up to 100 customers \citep{bogyrbayeva2021reinforcement}.

From the above, it can be seen that there are efficient and novel approaches in the state-of-the-art of EVRP. However, despite the effectiveness of these works, they still have the disadvantages of not being stable and robust when increasing the dimensionality of the problem. Generally, most of these methodologies are only used for EVs routing with a maximum of 100 to 320 clients. Therefore, in this article, we propose an intelligent and self-adaptive algorithm called  HHASA$_{RL}$ to solve CEVRP efficiently, obtaining optimal results at high-dimensional problems.

%% file: 4Background.tex
\section{Background}
\label{S:4}

This section provides background information of the essential building blocks for the HH proposed in this paper. First, it is detailed the HH by mentioning their importance, their classification, and the most commonly used techniques. Subsequently, the SA algorithm, as well as the Metropolis criterion, are presented. Then, the most used algorithms to deal with multi-armed bandit problems are detailed. Finally, the most commonly used improvement heuristics are shown.

\subsection{Hyper-heuristics}
HHs represent a class of high-level automated search techniques that aim to raise the level of generality and robustness with which search methods work to solve more complex problems. These algorithms explore a search space of low-level heuristics that can be neighborhood or movement operators, heuristic or metaheuristic algorithms. The two main categories of HH are heuristic generators and heuristic selectors. The generation of heuristics are methodologies used to create new heuristics from components of the existing ones. On the other hand, heuristics selection are methodologies to choose a heuristic among a set. The subject of this research will be the category of HH for selection, which controls a set of low-level heuristics during an iterative search process.

A generic heuristic selection consists of two key components, which are heuristic selection and move acceptance. As the name implies, the heuristic selection strategy must choose the most appropriate from a set of low-level heuristics at a certain point in the search process. While in the movement acceptance strategy, it is decided whether or not to accept the solution generated with the previous component. The most commonly used heuristic selection techniques are listed below. Among the most straightforward methods are random selection and random gradient, which consists of selecting a heuristic randomly and applying it iteratively until there is no improvement in fitness. Also, the greedy search uses all the perturbative heuristics from the available set and chooses the one with the best fitness \citep{pillay2018hyper}. On the other hand, strategies more commonly used by metaheuristic algorithms such as tournament selection and roulette wheel have also been used. In the roulette wheel strategy, each heuristic is associated with a probability calculated by dividing each score by the total score. Then a heuristic is randomly selected based on these probabilities. While tournament selection, a set of heuristics of fixed size is randomly selected to perform several tournaments, and the heuristic that solves with the best fitness is selected \citep{burke2013hyper}.

RL has also been used successfully, assigning scores to each heuristic within the available pool. A score is assigned to each heuristic with the RL methods based on its performance during the iterative process. During this process, by trial and error, the system tries to learn what heuristics to take by evaluating the status and accumulated rewards of the actions \citep{burke2013hyper}. In the same way, the most commonly used movement acceptance techniques are presented. The simplest strategy is to accept all moves regardless of the quality of the solutions, and another simple approach is to take motions that improve the solution's fitness. Local search techniques such as simulated annealing, late acceptance hill-climbing, and great deluge have also been used according to each of their specific strategies \citep{pillay2018hyper}.

\subsection{Simulated annealing}
\label{S:4SA}
This metaheuristic algorithm inspired by the physical process of annealing solid metals in metallurgy was proposed by Kirkpatrick et al. in 1983 to solve both global and combinatorial optimization problems \citep{kirkpatrick1983optimization}.

Taking the thermodynamic system as a reference, a candidate solution is generated in each iteration, considering improvement heuristics. The new candidate solution is accepted or rejected according to the Metropolis relation. This acceptance criterion is shown in Equation \ref{eq:pk} and is the key of the SA algorithm to avoid stagnation at local optima. The Metropolis criterion uses the relative quality of the solution and the temperature, which acts as a probability to select worse solutions, to improve the exploration of the search space \citep{delahaye2019simulated}.

\begin{equation}
    p^k=\left\lbrace
    \begin{array}{lll}
        \exp \left( \frac{-\Delta}{T} \right), & \textup{if} &\Delta > 0 \\
        1, & \textup{if} &\Delta \leq 0
    \end{array}
    \right.
    \label{eq:pk}
\end{equation}

The original pseudo-code of the SA is observed in Algorithm \ref{alg:SA}.

\begin{algorithm}
\caption{Pseudo-code of $SA$}
\label{alg:SA}
\begin{algorithmic}
\State \textbf{Inputs}: $I_{Iter}$, $\alpha$, $T_0$, and $M_{Acc}$
\State $\mathbf{s}$ $\gets$ Create initial solution
\State $T \gets T_0$
\While {$acc$ $<$ $M_{Acc}$}
    \For{(k $\gets$ 1 to $I_{Iter}$)}
        \State $\mathbf{s^\prime}$ $\gets$ Create neighbor solution($\mathbf{s}$)
        \State $\Delta = f(\mathbf{s^\prime})-f(\mathbf{s})$
        \If{$\Delta$ $\leq 0$}
            \State $\mathbf{s}$ $\gets$ $\mathbf{s^\prime}$
        \Else
            \If{$p^k > rand$}
                \State $\mathbf{s}$ $\gets$ $\mathbf{s^\prime}$
            \EndIf
        \EndIf
        \State$acc \gets acc+1$
        \State k $\gets$ k+1;
    \EndFor
    \State $T= T \cdot \alpha$
\EndWhile
\State \textbf{Return best solution found} 
\end{algorithmic}
\end{algorithm}

where the number of iterations for which the local search continues at a particular temperature is specified by $I_{Iter}$. At the same time, $\alpha$ is the coefficient controlling the cooling schedule, and $T_0$ is the initial temperature equal to the current temperature ($T$) at the start of the process. The $M_{Acc}$ represents the maximum number of function access allowed \citep{morales2019improved}.

\subsection{Multi-armed bandit RL methods}
\label{S:4Basic}

The multi-armed bandit is a classic RL problem whose name originates from a gambler sitting in front of a set of $n$ slot machines. The objective is to obtain the highest value of the accumulated reward, between each spin and using one machine at a time, whether to continue playing using the current machine or to switch to another one \citep{auer2002finite, gittins2011multi}. It optimizes its reward, acquiring knowledge (exploration) and optimizing decisions based on that learning (exploitation). Formally expressed and in a simple way, the multi-armed bandit problem of $K$-arms (actions or heuristics) is defined by $A_{i,n}$, which are random variables from $1\leq i \leq K$ and $n\geq 1$, where $i$ represents the index of the slot machine. There are many different solutions to tackle the multi-armed bandit problem. However, the most commonly used are  Epsilon Greedy ($epsilon$-$G$), Thompson Sampling ($TS$) and Upper Confidence Bound 1 ($UCB1$).

\subsubsection{Epsilon Greedy}
$\epsilon$-$G$ is a simple method for balancing exploration and exploitation by choosing between random exploration and exploitation. It is the greediest algorithm, as its name suggests, among the other two algorithms presented below. This is because the $\epsilon$ (valued between 0 and 1) refers to the probability of choosing to explore. However, the constant is usually set to 0.1, indicating that it exploits most of the time with a small possibility of exploring \citep{yang2021multi}. The pseudo-code of this strategy is presented in Algorithm \ref{alg:EpsilonGreedy}. In the vector \textit{\textbf{R}} are the accumulated rewards for each of the actions or heuristics until that moment.

\begin{algorithm}
\caption{Pseudo-code of $\epsilon$-$G$}
\label{alg:EpsilonGreedy}
\begin{algorithmic}
\State \textbf{Inputs}: $\epsilon$,  \textit{\textbf{R}}
\If {($rand > \epsilon$)}
    \State $heuristic$ $\gets$ Select a random action
\Else
    \State $heuristic$ $\gets$ Select the action with the $argmax$(\textit{\textbf{R}})
\EndIf
\State \textbf{Return} $heuristic$
\end{algorithmic}
\end{algorithm}

\subsubsection{Thompson Sampling}

On the other hand, $TS$ is an approach more based on Bayesian principles, which can produce more balanced and efficient results in some cases. A probability distribution (usually a beta distribution) of the actual success rate is constructed for each of the actions. This method uses information from the behavior of previously taken actions as training, creating an active exploration with a trial-and-error search of the behavior of the actions in each of the moves \citep{russo2017tutorial}. According to the heuristics results, it is decided whether to reward or punish that action by increasing the corresponding value of vector \textit{\textbf{R}} or vector \textit{\textbf{P}}, respectively. This strategy generates other actions which will maximize the reward, making future performance improvements and the pseudo-code is presented in Algorithm \ref{alg:ThompsonSampling}.

The greater the number of turns of the machine (action) that are completed successfully compared to failure, it means that a more significant number will be drawn, and the chances of that heuristic being chosen will increase. On the other hand, heuristics with a suboptimal success/failure rate will still have an opportunity to be selected, and in this way, the exploration is performed.

\begin{algorithm}
\caption{Pseudo-code of $TS$}
\label{alg:ThompsonSampling}
\begin{algorithmic}
\State \textbf{Inputs}: \textit{\textbf{R}}, \textit{\textbf{P}}
\For{(i $\gets$ 1 to $num_{actions}$)}
    \State $\theta_{i} \gets Beta(R_i+1, P_i+1)$ sample from Beta distribution
\EndFor
\State $heuristic$ $\gets$ Select the action with the $argmax(\pmb{\theta})$
\State \textbf{Return} $heuristic$
\end{algorithmic}
\end{algorithm}

\begin{algorithm}
\caption{Pseudo-code of $UCB1$}
\label{alg:UpperConfidenceBound}
\begin{algorithmic}
\State \textbf{Inputs}: $k$, \textit{\textbf{R}}, \textit{\textbf{S}}
\If{($k \leq num_{actions}$)}
    \State $heuristic \gets k$ 
\Else
    \For{(i $\gets$ 1 to $num_{actions}$)}
        \State $\phi_i$ $\gets R_i/{S_i}$ + $\frac{\sqrt{2 \cdot \log(k)}}{S_i}$
    \EndFor
\EndIf
\State $heuristic$ $\gets$ Select the action with the $argmax(\mathbf{\pmb{\phi}})$
\State $\mathit{S_{heuristic}}$  $\gets$ Is increased by 1
\State \textbf{Return} $heuristic$, \textit{\textbf{S}}
\end{algorithmic}
\end{algorithm}

\subsubsection{Upper Confidence Bound 1}

$UCB1$ is based on the principle of optimism in the face of uncertainty, which implies that if one is uncertain about an action, one optimistically assumes that the selected action is correct. Basically the main idea is to always choose a heuristic with the highest upper bound. Therefore, it uses the uncertainty in the estimation of the value of the machines to balance exploration and exploitation, which is one of the most critical characteristics of UCB1 \citep{umami2021comparing}.

Due to it explores further to systematically reduce uncertainty, its exploration decreases over time because it decays exponentially as the number of turns or shots of the machines increases. In other words, the least explored machine gets a boost even if its estimated average is low, especially if the gambler has been playing for a while. In this way, UCB1 can define its exploration-exploitation combinations without depending on any parameters. The pseudo-code is presented in Algorithm \ref{alg:UpperConfidenceBound}. The vector \textit{\textbf{R}} shows the rewards accumulated for each of the actions and the vector \textit{\textbf{S}} shows the number of times each of the heuristics has been selected.

\subsection{Heuristics for VRP}
\label{S:4Heu}
Heuristics seek to solve the problem based on the specific knowledge of the problem, usually suboptimal or close enough to a satisfactory solution. In the research field of VRP, heuristics are divided into constructive and improvement heuristics. Constructive heuristics are iterative methods used to generate initial solutions by constructing routes to which elements are added until the solution is complete. Solutions are constructed greedily and often produce solutions that are far from the optimal solution \citep{vidal2013heuristics}. The improvement heuristic explores the neighborhood of the current solution, searching for a better solution by applying perturbation operators. The local search stops when no improved solution is found in the neighborhood of the current solution, which is called a local optimum. The following are the most commonly used perturbation heuristics in the VRP literature:

\begin{itemize}
\item $Swap$: Two nodes are chosen randomly, either on the same route or on a different route, which exchange their position.
\item $Reversion$: Starting from two nodes, the number string is inverted regardless of whether they are not on the same route.
\item $2Opt$: Replaces two random arcs with two new ones with the possibility to reverse the direction of the route when reconnecting them.
\item $Insertion$: In the same way, two nodes are chosen, but the first node is inserted in the position after the second selected node.
\end{itemize}

%% file: 5Algorithm.tex
\section{Hyper-heuristic Adaptive Simulated Annealing and Reinforcement Learning (HHASA$_{RL}$)}
\label{S:5}

This section presents a detailed description of the HHASA$_{RL}$ algorithm. First, the blocks that make up the general flowchart of the HH are described. Following that, the local search block is explained, which is made up of four significant sub-blocks: $RL$ $Method$, $Generate$, $Repair$, and $Adjust Station$. Finally, the encoding used for this proposal is presented.

\subsection{General Description}

\begin{figure}[http]
\centering
\includegraphics[scale=0.20]{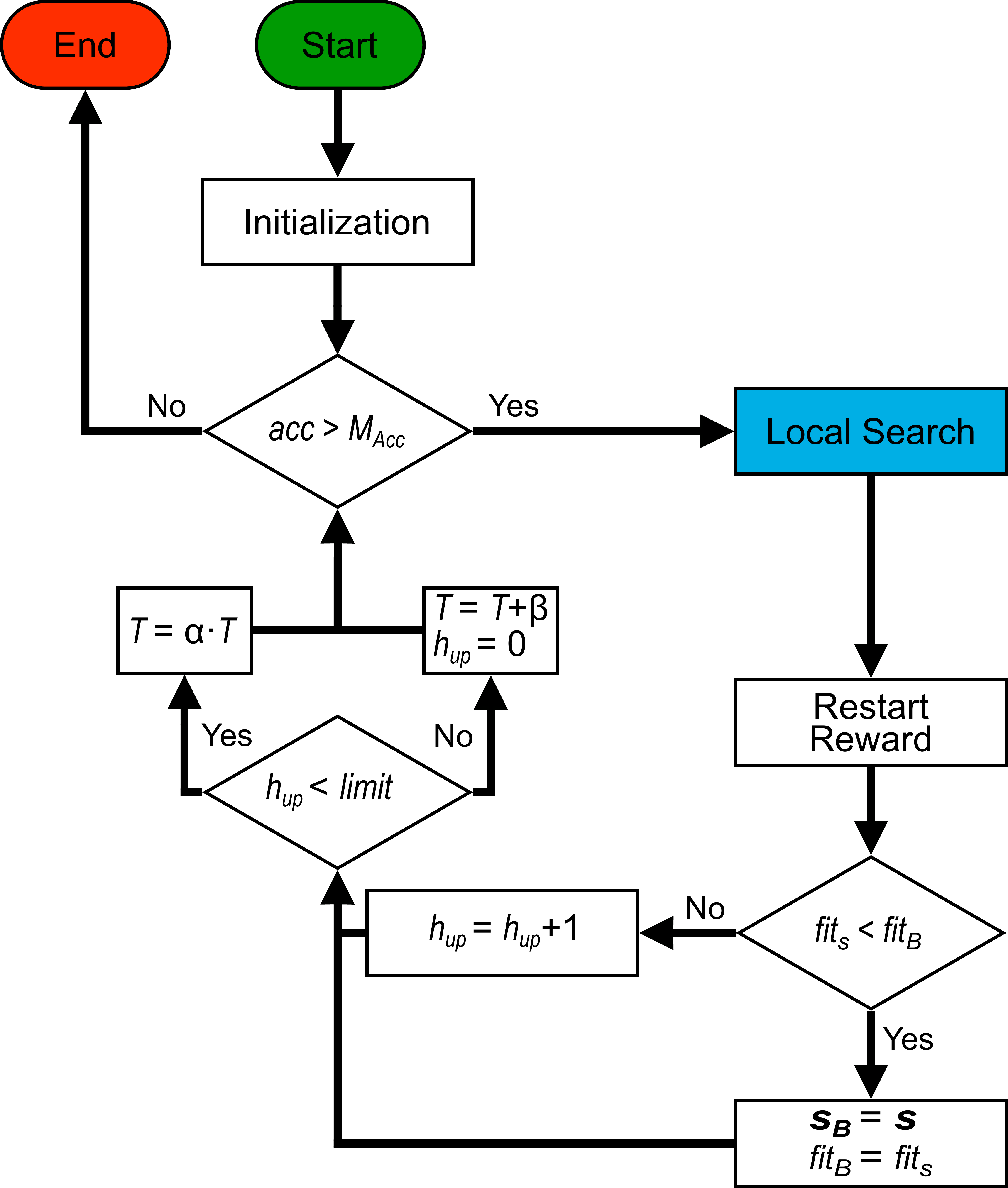}
\caption{The general flowchart of HHASA$_{RL}$}
\label{fig:flowchartGeneral}
\end{figure}

The general process of the HHASA$_{RL}$ algorithm is shown in Figure \ref{fig:flowchartGeneral}. The proposed method receives the same input parameters as the classical SA algorithm detailed in Section \ref{S:4SA}, which are $T$, $M_{Acc}$, $I_{Iter}$, $\alpha$. In addition to these parameters, $limit$ and the variable $h_{up}$ are used to regulate the temperature of the proposal. The parameter $limit$ indicates the maximum number of iterations that the variable $h_{up}$ can reach without improving the solution. If this condition is reached the T value will increase.

The Initialization procedure starts the internal variables of the HH and a feasible random solution ($\mathbf{s}$) is generated. This solution is obtained by randomly permuting all customers, separating them into routes for vehicles according to load, and adding stops at the nearest stations to recharge the battery when needed (before emptying the battery).

The generated solution $\mathbf{s}$ enters the loop where perturbations or local changes are performed until the maximum number of allowed fitness function accesses is reached (while $acc<M_{acc}$). The variable $acc$ monitors the number of evaluation function accesses used.

Then, the \textit{Local Search block} is in charge of performing $I_{Iter}$ perturbations to update the actual solution $\mathbf{s}$ and thus minimize the total distance of the tours. Next, the reward vector is restarted and the fitness of the best solution generated in the local search $fit_\mathbf{s}$ is compared with the fitness of the global best solution $fit_\mathbf{s_{B}}$. If the route distance is improved, the global solution is updated; however, if it is not improved, the value of $h_{up}$ is increased by 1. The variable $h_{up}$ is in charge of determining if the temperature of the algorithm should continue cooling or if it needs to be heated.

Lastly, if the variable $h_{up}$ is less than $limit$, it means that the fitness has improved steadily, so the temperature will continue to cool down and start again with the \textit{Local Search block} with the new temperature obtained through $T = \alpha \cdot T$. However, if the variable $h_{up}$ is equal to $limit$, it would mean that the solution has not improved, so the temperature will reheat by adding $\beta$ to increase the possibility of accepting worse solutions to escape from local minima through the Metropolis relation. The temperature $\beta$ to reheating is defined from the linear equation shown in Equation \ref{eq:beta}.

\begin{subequations}
\label{eq:beta}
\begin{align}
    \beta &= m \cdot (acc/M_{acc} \cdot 100)+(y_2-(m \cdot x_2))\\
    m &= (x_{max}-x_{min})/(y_{max}-y_{min})
\end{align}
\end{subequations}

where $x$ represent percentages of accesses to the objective function used. $x_{min}$ represents the minimum percentage and $x_{max}$ the maximum percentage by which the temperature will warm up if the fitness does not improve in $limit$ local searches.

On the other hand, the values of $y$ represent the temperatures that will be added to heat the system in the case of no improvement in fitness. Where $y_{min}$ represents the minimum temperature that will be increased to reheating the system and $y_{max}$ is the maximum temperature that is added.

\subsection{Local Search Block}

\begin{figure}[http]
\centering
\includegraphics[scale=0.20]{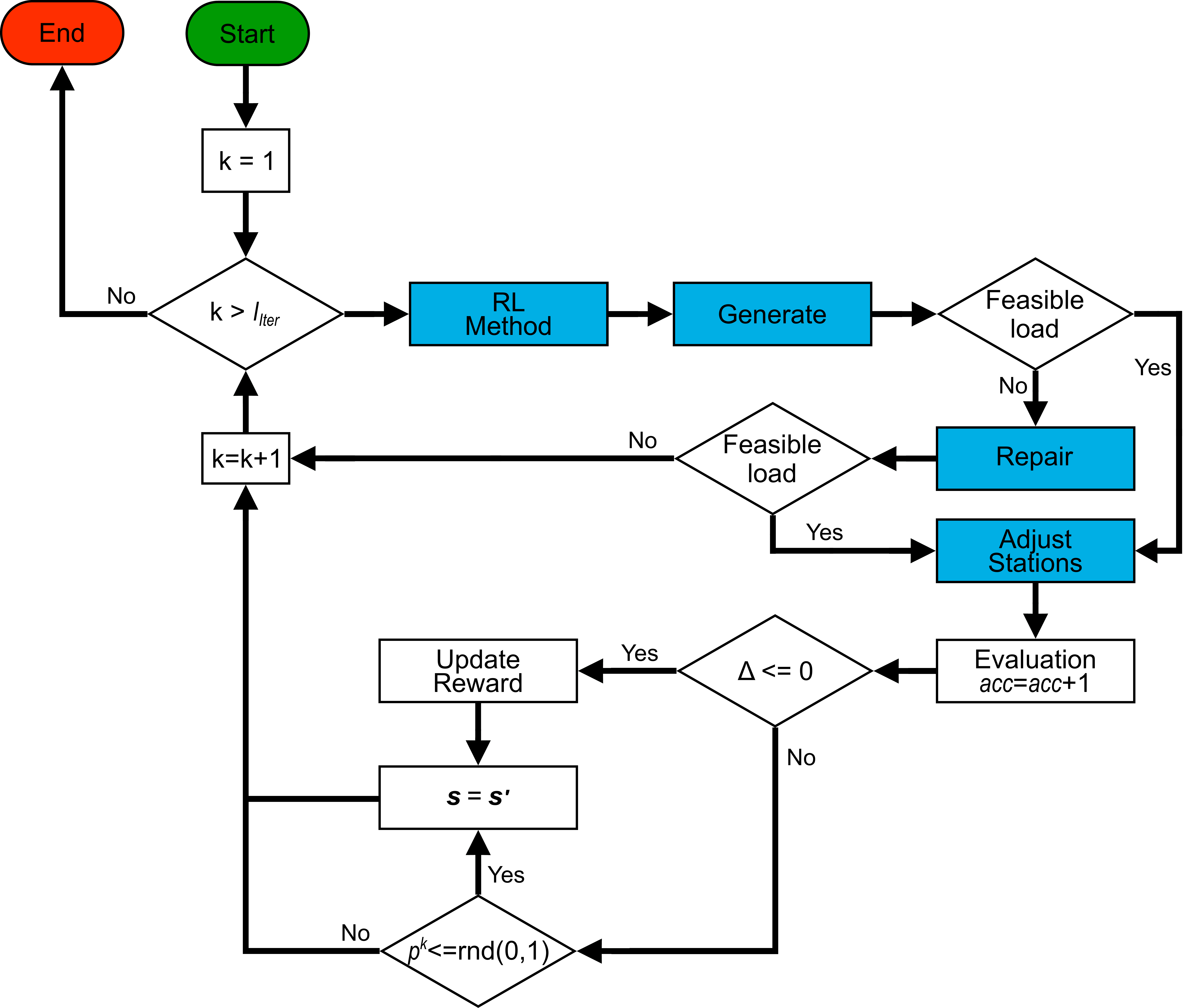}
\caption{Local search flowchart of HHASA$_{RL}$}
\label{fig:flowchartLocalSearch}
\end{figure}

Figure \ref{fig:flowchartLocalSearch} shows the flowchart of the local search of the proposed algorithm. The local search process begins with $I_{Iter}$ iterations at a constant temperature throughout the cycle. In the $RL$ $Method$ block, a multi-armed bandit RL method from those defined in Section \ref{S:4Basic} is used for the selection of the appropriate heuristic, based on the information contained in the reward vector for that specific local search. The most commonly used perturbation heuristics in the literature defined in Section \ref{S:4Heu} are used for the RL pool. This group includes the following eight heuristics: $Swap_{r1}$, $Reversion_{r1}$, $2Opt_{r1}$, $Insertion_{r1}$, $Swap_{r2}$, $Reversion_{r2}$, $2Opt_{r2}$ and $Insertion_{r2}$. The sub-index shown in each heuristic represents the closeness with which the customer $c_2$ will be selected to apply the heuristic regarding client $c_1$. The closeness values used can be a random selection among the $r1$\% or $r2$\% of customers closest to $c_1$ or among all customers.

The next step is to apply $Generate$ block to build a new solution and the pseudo-code of this block is shown in Algorithm \ref{alg:Generate}. This block applies the heuristic selected in the previous step on the $c_1$ and $c_2$ clients to modify the $\mathbf{s}$ solution. First of all, a random vector of all clients \textit{\textbf{cust}} is created to avoid repeating them when applying the heuristic. Customer $c_1$ is assigned and removed from the first element of the \textit{\textbf{cust}} vector. 

\begin{algorithm}
\caption{Pseudo-code of \textit{Generate block}}
\label{alg:Generate}
\begin{algorithmic}
\State \textbf{Inputs}: $\mathbf{s}$, $heuristic$, $n_c$ and \textit{\textbf{cust}}
\If {$\sim$ isempty({\textbf{cust}})}
    \State \textit{\textbf{cust}} $\gets$ permutation($n_c$)
\EndIf
\State $c_1 \gets cust(1)$
\State Remove $c_1$ from \textit{\textbf{cust}}
\State Select $c_2$
\State $\mathbf{s^\prime}$ $\gets$ Apply the $heuristic$ to  $c_1$ and $c_2$
\State $\mathbf{s^\prime}$ $\gets$ Eliminate stations that are next to each other
\State \textbf{Return} $\mathbf{s^\prime}$ and \textit{\textbf{cust}}
\end{algorithmic}
\end{algorithm}

The constraint of the load capacity is checked to the newly generated solution $\mathbf{s^\prime}$. If the load constraint is not satisfied, then the $Repair$ block is used to adjust the generated path and make the EV loads feasible. The pseudo-code for this block is presented in Algorithm \ref{alg:Repair} and the following explains how it works. First, the last clients that exceed EV charging are stored in the vector \textit{\textbf{repair}}. Then, the positions of the customers on other routes where it would be feasible to add the load without exceeding it are obtained from each of the customers of \textit{\textbf{repair}}.

There are two possibilities when repairing the loads; the first one is that there is capacity in other EVs to add these loads, so in this case, it is added in the position to one side of the customer with the one that presents the minimum distances. On the other hand, the opposite case is that there is no free space in the other vehicles to add more charges. So, if it cannot be repaired, one is added to the variable of $k$, and the steps mentioned above will be repeated starting from the selection of the heuristic in the $RL$ $Method$ block.

Later, in cases where $\mathbf {s^\prime}$ is feasible, the \textit{Adjust Station block} will be applied. In this block, depending on a probability, a charging station can be moved or removed from the route as shown in Algorithm \ref{alg:Adjust} through the roulette wheel selection. The probabilities for the actions mentioned in this block are $p_m$ that is the percentage for move and $p_e$ for eliminate. However, it is important to highlight that the stations will be added to satisfy the power restriction, ensuring that electric vehicles do not run out of battery during the journey.

\begin{algorithm}
\caption{Pseudo-code of $Repair$ block}
\label{alg:Repair}
\begin{algorithmic}
\State \textbf{Inputs}: $\mathbf{s^\prime}$
\For{(v $\gets$ 1 to $num_{routes}$)}
    \State $q_T$(v) $\gets$ Check load of the total route(v)
    \If {($q_T(v) > Max_C$)}
        \State \textit{\textbf{repair}} $\gets$ Last customer exceeding $Max_C$
    \EndIf
\EndFor
\State $\mathbf{s^\prime}$ $\gets$ Move \textit{\textbf{repair}} next to the customer with the minimum distance of the feasible routes to add the load

\State \textbf{Return} $\mathbf{s^\prime}$
\end{algorithmic}
\end{algorithm}

Afterward, an evaluation of the objective function is performed to obtain the total sum of the distance of the vehicles; to subsequently calculate $\Delta$ and decide whether to reward the selected heuristic for iteration $k$ and update the best solution or apply the Metropolis relation to accept movements with worse fitness. This local search continues to iterate until $k$ reaches the number $I_{Iter}$.

\begin{algorithm}
\caption{Pseudo-code of $Adjust Station$ block}
\label{alg:Adjust}
\begin{algorithmic}
\State \textbf{Inputs}: $\mathbf{s^\prime}$ and $memory$
\For{(v $\gets$ 1 to $num_{routes}$)}
    \State $e_T$(v) $\gets$ Check the total energy of the route(v)
    \If {($e_T(v) > Max_Q$)}
        \State $\mathbf{s^\prime}$ $\gets$ Add station where needed
    \Else
        \If {($rand > memory$)}
            \State $ii \gets$ Roulette Wheel Selection ({Move, Eliminate})
            \State $\mathbf{s^\prime}$ $\gets$ Apply($ii$)
        \EndIf 
    \EndIf 
\EndFor
\State \textbf{Return} $\mathbf{s^\prime}$
\end{algorithmic}
\end{algorithm}

\subsection{Solution Encoding}

The encoding used by the proposed algorithm is as follows. The depot is represented by the number 0, while the numbers from 1 to $n_c + n_s$ uniquely represent both route customers or charging stations. The EV routes are separated by 0s because each of them has to start and end at the depot. The encoding for adding battery recharging stations is represented by the number -1. An example of this representation can be seen in Figure \ref{fig:evrp}, that the encoding for EV A would be [0, 1, -1, 2, 3, 4, -1, 5, 0]. The -1 representing the battery recharge is replaced by the index of the nearest station between the two clients. So the complete example of the four routes presented in Figure \ref{fig:evrp} would be as follows: [0, 1, \underline{16}, 2, 3, 4, \underline{16}, 5, 0, 6, 7, 0, 8, 15, \underline{18}, 14, 0, 9, \underline{19}, 10, 11, \underline{20}, 12, 13, 0], where the underlined numbers are the charging stations.

%% file: 6Experimental.tex
\section{Experimental Framework}
\label{S:6}

In this section, we present detailed information of the benchmark used. Also, the algorithms used for comparison, statistical and non-parametric tests used to determine the performance and efficiency of the proposed algorithm are established.

\subsection{Benchmark description}

The recent publicly available benchmark proposed for the IEEE WCCI2020 competition on computational intelligence for CEVRP is used to test the performance of the proposed approach \citep{mavrovouniotis2020benchmark}. This benchmark contains 17 instances, of which seven are small with up to 100 customers and ten are large with problems of up to 1000 customers. The detailed information of each of the cases is summarized in Table \ref{Tab:Details}.

\begin{table}[H]
    \caption{Details of the CEVRP benchmark set}
    \label{Tab:Details}
    \centering
    \begin{tabular}{*7c}
        \hline
        \textbf{Name} & \textbf{Customers ($n_c$)} & \textbf{Stations ($n_s$)} & \textbf{$Min_{Routes}$} & \textbf{$Max_C$} & \textbf{$Max_Q$} & \textbf{$h$} \\
        \hline
        E22 & 21 & 8 & 4 & 6000 & 94 & 1.2 \\
        E23 & 22 & 9 & 3 & 4500 & 190 & 1.2 \\
        E30 & 29 & 6 & 4 & 4500 & 178 & 1.2 \\
        E33 & 32 & 6 & 4 & 8000 & 209 & 1.2 \\
        E51 & 50 & 5 & 5 & 160 & 105 & 1.2 \\
        E76 & 75 & 7 & 7 & 220 & 98 & 1.2 \\
        E101 & 75 & 9 & 8 & 200 & 103 & 1.2 \\
        \hline
        X143 & 142 & 4 & 7 & 1190 & 2243 & 1.0 \\
        X214 & 213 & 9 & 11 & 944 & 987 & 1.0 \\
        X352 & 351 & 35 & 40 & 436 & 649 & 1.0 \\
        X459 & 458 & 20 & 26 & 1106 & 929 & 1.0 \\
        X573 & 572 & 6 & 30 & 210 & 1691 & 1.0 \\
        X685 & 684 & 25 & 75 & 408 & 911 & 1.0 \\
        X749 & 748 & 30 & 98 & 396 & 790 & 1.0 \\
        X819 & 818 & 25 & 171 & 358 & 926 & 1.0 \\
        X916 & 915 & 9 & 207 & 33 & 1591 & 1.0 \\
        X1001 & 1000 & 9 & 43 & 131 & 1684 & 1.0 \\
        \hline
        \hline
    \end{tabular}
\end{table}

The columns of the table present the number of customers, the number of charging stations distributed in the space, the minimum number of EVs ($Min_{Routes}$), the maximum load of an EV ($Max_C$), the battery charge of an EV ($Max_Q$) and the energy consumption constant ($h$). It is important to mention that there is only one depot from which all EVs depart and return at the end of the route. The main objective of this work is to reduce the total distance that vehicles have to travel. So it is possible to have a solution that consists of more EVs.

\subsubsection{Baseline algorithms and compared methods}

For the comparisons, the three winning algorithms of the CEVRP competition at the IEEE WCCI2020 conference, which are named VNS, SA and GA, are considered. Also, it is compared with the proposal of Woller et al. called Greedy Randomized Adaptive Search Procedure (GRASP) \citep{woller2020grasp}. Furthermore, a comparison is also made with the results of Jia et al. from 2021. The authors proposed an improved algorithm called Bilevel Ant Colony Optimization (BACO), which presents the best state-of-the-art results on this benchmark.

On the other hand, three versions of the proposed algorithm are compared using the simple methods for solving the multi-armed bandit problem, named HHASA$_{\epsilon-G}$, HHASA$_{UCB_1}$ and HHASA$_{TS}$. In addition, to evaluate the improvement of simple RL algorithms, a version that replaces the RL block with a random selection of heuristics called HHASA is used.

\subsection{Experimental set-up}

In order to provide a fair comparison, the proposal followed all the evaluation criteria mentioned in the competition. These are 20 independent runs with random seeds and a stopping criterion of $25.000\cdot n_c$ maximum evaluations to the objective function ($M_{Acc}$), where $n_c$ is the instance size of the problem. Also, the set values of the internal parameters of the algorithm are $\alpha$ of 0.99, $limit$ of 20, $I_{iter}$ of $40\cdot n_c$, $60$\% of $p_m$ and $40$\% of $p_e$. The values of the parameters used to obtain the $\beta$ value in the reheating stage are 0\% and 90\% for $x_{min}$ and $x_{max}$ and 0.05 and 1 for $y_{min}$ and $y_ {max}$, respectively. Finally, for the heuristics, a value of 10\% is selected for r1 and 100\% for r2. All experiments were done using Matlab 9.4 on an Intel Core i5 CPU @ 2.7Ghz with 16 GB of RAM.

Since the codes of the proposed algorithms are not available for comparison to determine the efficiency of the proposed method, the results reported in the literature are used directly, which are the minimum (min), mean and standard deviation (std) values over 20 runs. In addition, to assess that the statistical differences observed among the performance of the algorithms are statistically significant, we use the Friedman's non-parametric test for multiple comparisons \citep{scoczynski2021saving, mousavirad2022population} and the Holm's poct-hoc test for 1-to-n comparisons \citep{aziz2016estimation}.

%% file: 7Results.tex
\section{Results and Discussion}
\label{S:7}
This section provides experimental results of the different aspects used to evaluate the efficiency of the HH proposal. First, a comparison is made with the statistical results of the independent runs of the proposals with the state-of-the-art. Then, the non-parametric analysis is showed to rank the algorithms and determine if there is a significant difference in their mean. Subsequently, a visual comparison of the selection of heuristics among the multi-armed bandit RL methods is presented. Next, a comparison of the difference between the mean of the solutions against the best fitness found is reported. Finally, a visualization of the solution generated for four instances is displayed.

\subsection{Comparison of the statistical results}

The statistical results obtained by applying all instances of the CEVRP benchmark in the algorithms mentioned in the previous section are shown in Tables \ref{Tab:Small} and \ref{Tab:Large}. These tables present the min, mean and std values of the solutions obtained by the algorithms in the 20 independent runs. It is important to mention that the minimum average distances are indicated in bold and with an asterisk. While the results presented only in bold are those that beat the minimum mean obtained by the state-of-the-art. 

According to the short instances in Table \ref{Tab:Small}, the following observations can be obtained. According to the statistical test, the four variants of the proposed HH only present worse average results than BACO in instances E51 and E76. While for the other cases, HHASA$_{RL}$ and HHASA have equivalent performance with the compared algorithms and are superior for E33 and E101. On the other hand, the HHASA$_{UCB1}$ algorithm found the smallest distance values for the objective function in the seven small instances. HHASA$_{\epsilon-G}$ found the best fitness in six cases with E101 missing, while HHASA$_{TS}$ and HHASA$_{TS}$ proposals found five. In general, the results reveal that the HHASA$_{UCB1}$ algorithm is very effective in finding the lowest fitness. However, it is difficult to determine which algorithm is better and presents better robustness in these types of instances because HHASA$_{UCB1}$, HHASA$_{TS}$ and BACO present better mean results in five cases.

\begin{table}[htbp]
    \caption{Results of the proposed algorithm applied to small instances of the benchmark}
    \label{Tab:Small}
    \centering
    \resizebox{18cm}{!} {
    \begin{tabular}{cc|cccc|ccccc}\midrule
    \textbf{Instances} & \textbf{Values} & \textbf{HHASA$_{TS}$} & \textbf{HHASA$_{UCB_1}$} & \textbf{HHASA$_{\epsilon-G}$} & \textbf{HHASA} & \textbf{BACO} & \textbf{VNS} & \textbf{SA} & \textbf{GA} & \textbf{GRASP} \\
    \midrule
    \multirow{3}{*}{E22} & min & 384.67 & 384.67 & 384.67 & 384.67 & 384.67 & 384.67 & 384.67 & 384.67 & 389.82 \\
    & mean & \textbf{384.67}* & \textbf{384.67}* & \textbf{384.67}* & \textbf{384.67}* & \textbf{384.67}* & \textbf{384.67}* & \textbf{384.67}* & \textbf{384.67}* & 389.89 \\
    & std & 0 & 0 & 0 & 0 & 0 & 0 & 0 & 0 & 0.41\\
    \midrule   
    \multirow{3}{*}{E23} & min & 571.94 & 571.94 & 571.94 & 571.94 & 571.94 & 571.94 & 571.94 & 571.94 & 571.94\\
     & mean & \textbf{571.94}* & \textbf{571.94}* & \textbf{571.94}* & 572.51 & \textbf{571.94}* & \textbf{571.94}* & \textbf{571.94}* & \textbf{571.94}* & 572.36 \\
     & std & 0 & 0 & 0 & 2.54 & 0 & 0 & 0 & 0 & 0.56 \\
    \midrule   
    \multirow{3}{*}{E30} & min & 509.47	& 509.47 & 509.47 & 509.47 & 509.47 & 509.47 & 509.47 & 509.47 & 512.19 \\
     & mean & \textbf{509.47}* & \textbf{509.47}* & \textbf{509.47}* & \textbf{509.47}* & \textbf{509.47}* & \textbf{509.47}* & \textbf{509.47}* & \textbf{509.47}* & 512.67 \\
     & std & 0 & 0 & 0 & 0 & 0 & 0 & 0 & 0 & 0.31 \\
    \midrule   
    \multirow{3}{*}{E33} & min & 840.14	& 840.14 & 840.14 & 840.14 & 840.57 & 840.14 & 840.57 & 844.25 & 841.08 \\
     & mean & 840.70 & \textbf{840.41}* & 840.82 & 841.10 & 842.30 & 840.43 & 854.07 & 845.62 & 845.06 \\
     & std & 1.40 & 0.57 & 1.28	& 2.95 & 1.42 & 1.18 & 12.80 & 0.92 & 1.56  \\
    \midrule   
    \multirow{3}{*}{E51} & min & 529.90	& 529.90 & 529.90 & 529.90 & 529.90 & 529.90 & 533.66 & 529.90 & 536.09\\
     & mean & 536.98 & 535.13 &	534.16 & 535.09 & \textbf{529.90}* & 543.26 & 533.66 & 542.08 & 546.21\\
     & std & 7.27 & 7.25 & 7.95 & 7.30 & 0 & 3.52 & 0 & 8.57 & 5.32\\
    \midrule
    \multirow{3}{*}{E76} & min & 692.74 & 692.64 & 692.64 & 694.54 & 692.64 & 692.64 & 701.03 & 697.27 & 701.63\\
     & mean &694.96	& 695.65 & 695.14 & 694.94 & \textbf{692.85}* & 697.89 & 712.17 & 717.30 & 711.36\\
     & std &1.63 & 2.47 & 2.84 & 1.02 & 0.81 & 3.09 & 5.78 & 9.58 & 5.27\\
    \midrule   
    \multirow{3}{*}{E101} & min & 837.10 & 835.63	& 836.17 & 837.10 & 840.25 & 839.29 & 845.84 & 852.69 & 847.47\\
     & mean & \textbf{843.10}* & \textbf{843.31} & \textbf{844.77} & \textbf{843.17} & 845.95 & 853.34 & 852.48 & 872.69 & 856.86\\
     & std & 3.90 & 4.04 & 5.61 & 3.79 & 4.58 & 4.73 & 3.44 & 9.58 & 6.90\\
    \midrule   
    \end{tabular}
    }
\end{table}

Subsequently, the algorithms are compared on large instances in Table \ref{Tab:Large}. From these results, the following insights can be obtained. According to the results, the four variants of the proposed HH present a lower mean value than the state-of-the-art results for the largest instances, which are X573, X685, X749, X819, X916 and X1001. In addition, the proposed HHASA$_{TS}$ also outperforms the X214 case, so it has found the best average fitness in six out of ten cases with a large number of customers, being the best variant within the proposed HHs. On the other hand, at the minimum values found for the objective function, the HHASA$_{TS}$ algorithm updated the best-known solutions at five instances out of ten, that are X214, X685, X819, X916 and X1001. While the HHASA$_{UCB1}$ algorithm updated the fitness at X573 with a very small difference of 0.90 with respect to the best distance found by HHASA$_{TS}$. Also, the HHASA$_{\epsilon-G}$ updated the best fitness for the X459 instance; and finally, the HHASA the X749. Overall, the results reveal that for large customer instances, the HHASA$_{TS}$ approach is effective in finding the best mean values over the 20 independently runs and has updated the minimum best-known solutions. 

\begin{table}[htbp]
    \caption{Results of the proposed algorithm applied to large instances of the benchmark}
    \label{Tab:Large}
    \centering
    \resizebox{18cm}{!} {
    \begin{tabular}{cc|cccc|ccccc}\midrule
    \textbf{Instances} & \textbf{Values} & \textbf{HHASA$_{TS}$} & \textbf{HHASA$_{UCB_1}$} & \textbf{HHASA$_{\epsilon-G}$} & \textbf{HHASA} & \textbf{BACO} & \textbf{VNS} & \textbf{SA} & \textbf{GA} & \textbf{GRASP} \\
    \midrule
    \multirow{3}{*}{X143} & min & 15910.86 & 15912.77 & 15899.86 & 15921.68 & 15901.23 & 16028.05 & 16610.37 & 16488.60 & 16460.80\\
     & mean & 16214.37 & 16231.33 & 16173.06 & 16271.78 & \textbf{16031.46}* & 16459.31 & 17188.90 & 16911.50 & 16823.00\\
     & std & 215.77	& 173.73 & 198.91 & 250.09 & 262.47 & 242.59 & 170.44 & 282.30 & 157.00\\
    \midrule   
    \multirow{3}{*}{X214} & min & 11090.28 & 11097.63 & 11098.34 & 11120.28 & 11133.14 & 11323.56 & 11404.44 & 11762.07 & 11575.60\\
     & mean & \textbf{11206.60}* & 11260.83 & 11247.30 & 11251.80 & 11219.70 & 11482.20 & 11680.35 & 12007.06 & 11740.70\\
     & std & 84.58 & 88.73 & 99.53 & 73.03 & 46.25 & 76.14 & 116.47 & 156.69 & 80.41\\
    \midrule
    \multirow{3}{*}{X352} & min & 26622.42 & 26549.88 & 26486.05 & 26606.06 & 26478.34 & 27064.88 & 27222.96 & 28008.09 & 27521.20\\
     & mean & 26750.60 & 26760.35 &	26760.58 & 26812.89 & \textbf{26593.18}* & 27217.77 & 27498.03 & 28336.07 & 27775.30\\
     & std & 102.55	& 116.44 & 135.77 & 90.44 & 72.86 & 86.20 & 155.62 & 205.29 & 111.99\\
    \midrule 
    \multirow{3}{*}{X459} & min & 24794.35 & 24769.67 & 24752.03 & 24815.37
    & 24763.93 & 25370.80 & 27222.96 & 26048.21 & 25929.20\\
     & mean & 25041.10 & 25036.67 &	24979.89 & 25060.02 & \textbf{24916.60}* & 25582.27 & 25809.47 & 26345.12 & 26263.30\\
     & std & 237.58	& 114.68 & 151.54 & 121.58 & 94.08 & 106.89 & 157.97 & 185.14 & 134.66\\
    \midrule
    \multirow{3}{*}{X573} & min & 51436.90 & 51436.00 & 51485.68 & 51545.10 & 53822.87 & 52181.51 & 51929.24 & 54189.62 & 52584.50\\
     & mean & \textbf{51776.70} & \textbf{51764.24} &	\textbf{51771.50} & \textbf{51748.42}* & 54567.15 & 52548.09 & 52793.66 & 55327.62 & 52990.90\\
     & std & 166.86	& 152.69 & 158.01 & 119.57 & 231.05 & 278.85 & 577.24 & 548.05 & 246.79\\
    \midrule
    \multirow{3}{*}{X685} & min & 69955.95 & 70348.53 & 70323.62 & 70413.81 & 70834.88 & 71345.40 & 72549.90 & 73925.56 & 72481.60\\
     & mean & \textbf{70401.25}* & \textbf{70719.10} & \textbf{70684.34} & \textbf{70791.10} & 71440.57 & 71770.57 & 73124.98 & 74508.03 & 72792.70\\
     & std & 218.98	& 291.53 & 174.26 &	222.85 & 281.78 & 197.08 & 320.07 & 409.43 & 189.53\\
    \midrule
    \multirow{3}{*}{X749} & min & 79779.87 & 79829.23 & 79850.73 & 79732.99 & 80299.76 & 81002.01 & 81392.78 & 84034.73 & 82187.30\\
     & mean & \textbf{80135.67}* & \textbf{80256.36} & \textbf{80318.42} & \textbf{80397.82} & 80694.54 & 81327.39 & 81848.13 & 84759.79 & 82733.40\\
     & std & 219.50	& 303.30 & 399.47 & 432.11 & 223.91 & 176.19 & 275.26 & 376.10 & 213.21\\
    \midrule
    \multirow{3}{*}{X819} & min & 161924.79 & 162350.42 & 162387.34 & 162523.88 & 164720.80 & 164289.95 & 165069.77 & 170965.68 & 166500.00\\
     & mean & \textbf{162530.67}* & \textbf{162819.78} & \textbf{162883.17} & \textbf{163031.19} & 165565.79 & 164926.41 & 165895.78 & 172410.12 & 166970.00\\
     & std & 289.41	& 258.35 & 300.11 & 389.12 & 401.02 & 318.62 & 403.70 & 568.58 & 211.84\\
    \midrule
    \multirow{3}{*}{X916} & min & 336717.71 & 337200.96 & 337520.94 & 338007.56 & 342993.01 & 341649.91 & 342796.88 & 357391.57 & 345777.00\\
     & mean & \textbf{337641.92}* & \textbf{338349.57} & \textbf{338639.53} & \textbf{338688.50} & 344999.95 & 342460.70 & 343533.85 & 360269.94 & 347269.00\\
     & std & 461.47	& 454.28 & 544.69 & 328.64 & 905.72 & 510.66 & 556.98 & 229.19 & 654.93\\
    \midrule
    \multirow{3}{*}{X1001} & min & 75469.29 & 75864.07 & 75782.95 & 75850.15 & 76297.09 & 77476.36 & 78053.86 & 78832.90 & 77636.20\\
     & mean & \textbf{75931.28}* & \textbf{76131.56} & \textbf{76245.73} & \textbf{76234.51} & 77434.33 & 77920.52 & NA & 79163.34 & 78111.20\\
     & std & 304.10	& 212.24 & 226.30 & 271.18 & 719.86 & 234.73 & 306.27 & NA & 315.31\\
    \midrule   
    \end{tabular}
    }
\end{table}

\subsection{Non-parametric analysis}

The results of the non-parametric analysis obtained through the Friedman test for multiple comparisons and the Post Hoc Holm's test for 1-to-n comparisons taking into account the 17 instances of the benchmark are shown in Table \ref{Tab:HHFriedman}. The p-values of the Friedman's test with a value lower than 0.05 are presented in bold, which means that the null hypothesis of equal performance can be discarded. Furthermore, the HHASA$_{TS}$ algorithm has the best ranking according to the Friedman test. It is also important to note that the three HH proposals using multi-armed bandit RL methods present a better ranking than the HHASA with its random heuristic selection which means that the RL methods are beneficial. Besides, the $p_{Holm}$ of the HHASA is lower than 0.05, which indicates that its performance is significantly worse compared to the control algorithm that is HHASA$_{TS}$.

\begin{table}[!htp]
\centering
\caption{Average Friedman’s rankings and Holm’s $p$ values (0.05) of the four proposed HHs for the CEVRP benchmark.}
\label{Tab:HHFriedman}
    \begin{tabular}{l|c|c} \midrule
    Algorithm & Ranking & $p_{Holm}$ \\ \midrule
    HHASA$_{TS}$ & 1.8824  &  \\
    HHASA$_{UCB1}$ & 2.4706 & 0.287886 \\
    HHASA$_{\epsilon-G}$ & 2.5294 & 0.287886 \\
    HHASA & 3.1176 & \textbf{0.015828} \\
    \midrule
    \end{tabular}
\end{table}

Another non-parametric analysis is shown below, but now it is with the three HH proposals containing the RL block and the five state-of-the-art algorithms in Table \ref{Tab:AllFriedman}. This table presents two comparisons; in the first three columns using all instances and the remaining columns using 11 large instances starting from E101 to X1001. In this analysis, it can be identified that for both comparisons, the first three ranking places are held by the HHs proposals of this work, being HHASA$_{TS}$ the best ranked. Also, according to $p_{Holm}$ using HHASA$_{TS}$ as a control, there is a significant difference for the VNS, SA, GRASP and GA algorithms. Finally, it is possible to observe that although there is no significant difference with the BACO algorithm, the performance of the HHASA$_{TS}$ algorithm is superior in the average results and in the ranking.

\begin{table}[!htp]
\centering
\caption{Average Friedman’s rankings and Holm’s $p$ values (0.05) of the comparison with the state-of-the-art for the CEVRP benchmark.}
\label{Tab:AllFriedman}
\begin{tabular}{l|c|c|c|c} \midrule
Algorithm & Ranking $_{all}$ & $p_{Holm}$ $_{all}$ & Ranking $_{\geq E101}$ & $p_{Holm}$ $_{\geq E101}$ \\ \midrule
HHASA$_{TS}$ & 2.4118 & & 1.7273 & \\
HHASA$_{UCB1}$ & 2.8824 & 0.882417 & 2.5455 & 0.676703\\
HHASA$_{\epsilon-G}$ & 3.0588 & 0.882417 & 2.7273 & 0.676703\\
BACO & 3.4118 & 0.701859 & 3.5455 & 0.245168\\
VNS & 4.6471 & \textbf{0.031207} & 4.8182 & \textbf{0.012333}\\
SA & 5.6471 & \textbf{0.000589} & 6.0909 & \textbf{0.000147}\\
GRASP & 6.8824 & \textbf{0.000001} & 6.6364 & \textbf{0.000016}\\
GA & 7.0588 & \textbf{0} & 7.9091 & \textbf{0}\\
\midrule
\end{tabular}
\end{table}

\subsection{Analysis of the selection of heuristics of the HH proposals}

\begin{figure}[H] 
\centering
    \begin{subfigure}{0.11\textwidth}
        \begin{subfigure}{\textwidth}
        \centering
        \includegraphics[width=\linewidth]{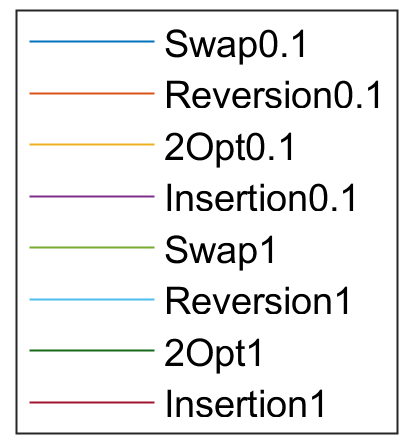}
        \end{subfigure}
    \end{subfigure}
    \begin{subfigure}{0.2\textwidth}
        \begin{subfigure}{1.10\textwidth}
        \centering
        \includegraphics[width=\linewidth]{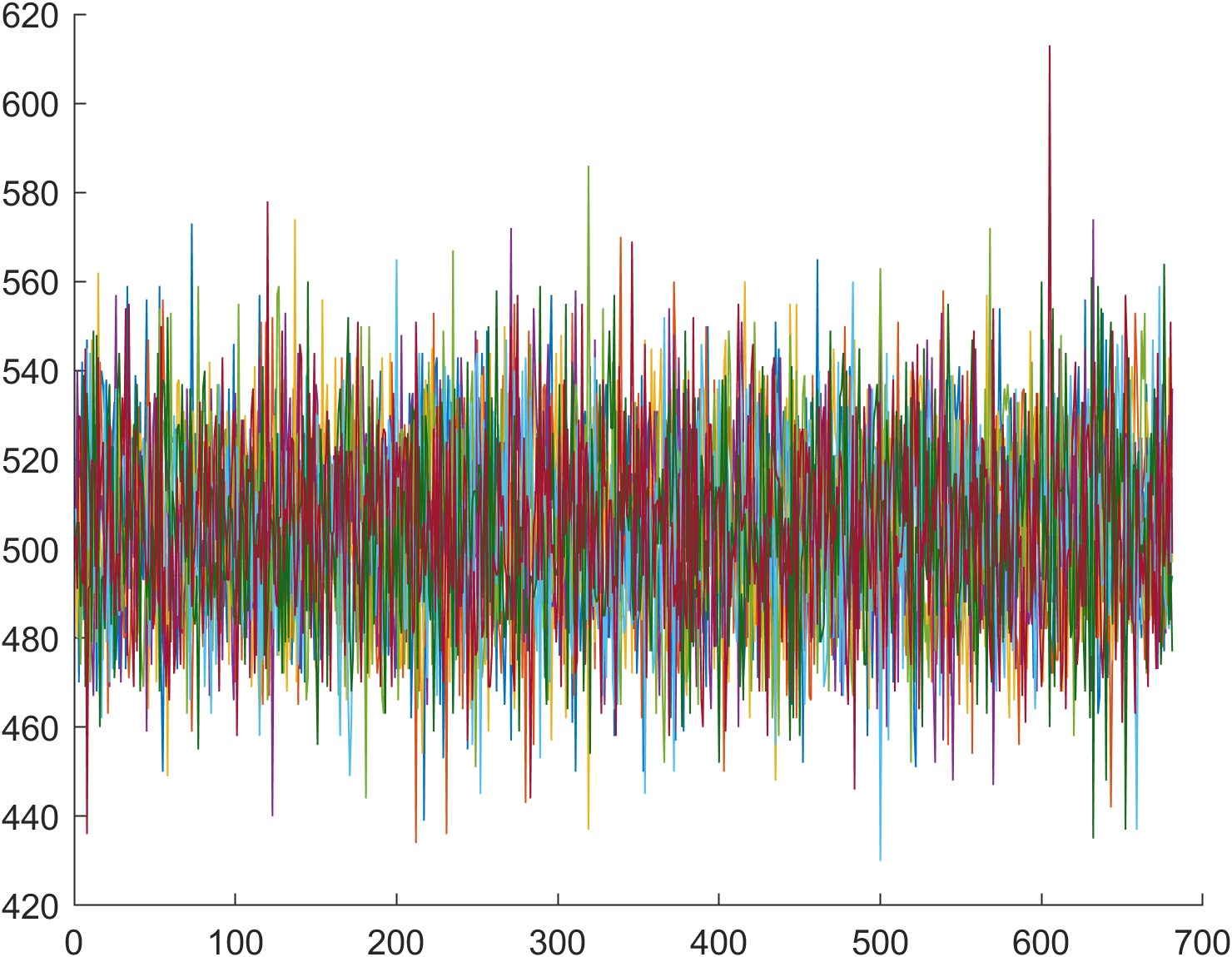}
        \end{subfigure}
        \begin{subfigure}{1.10\textwidth}
        \centering
        \includegraphics[width=\linewidth]{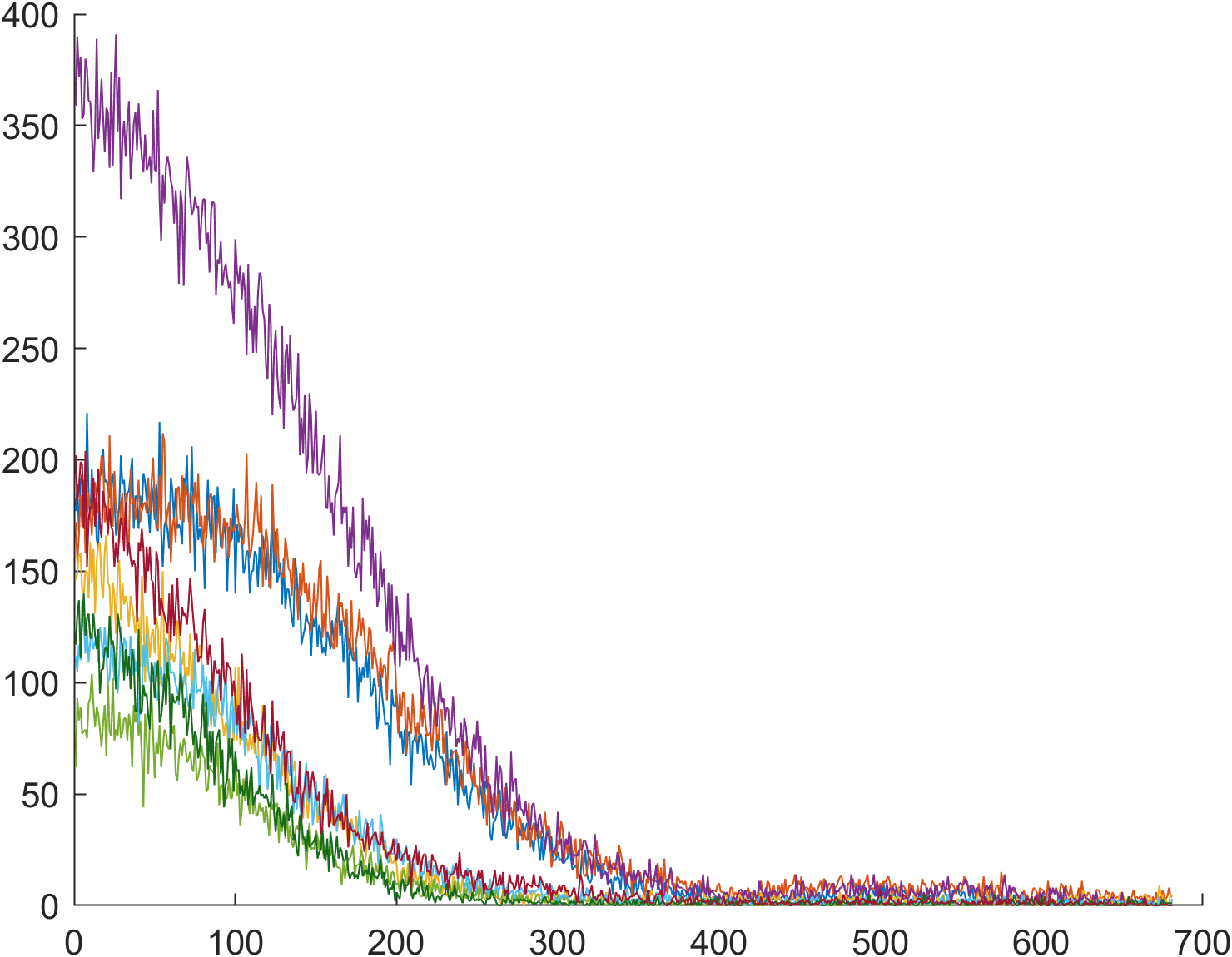}
        \end{subfigure}
    \caption{HHASA}
    \label{fig:E101Rand}
    \end{subfigure}
        \hspace{0.2cm}
    \begin{subfigure}{0.2\textwidth}
        \begin{subfigure}{1.10\textwidth}
        \centering
        \includegraphics[width=\linewidth]{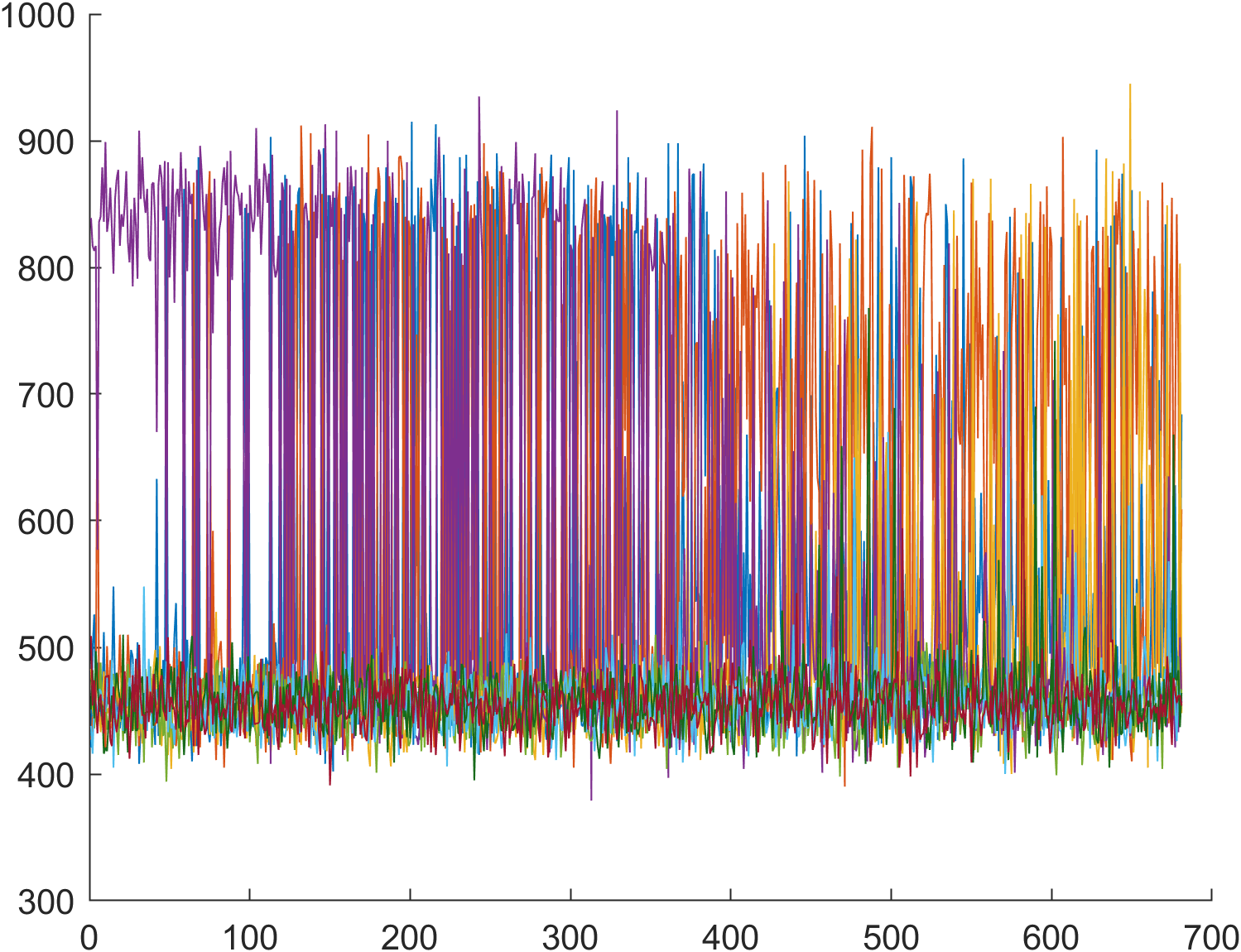}
        \end{subfigure}
        \begin{subfigure}{1.10\textwidth}
        \centering
        \includegraphics[width=\linewidth]{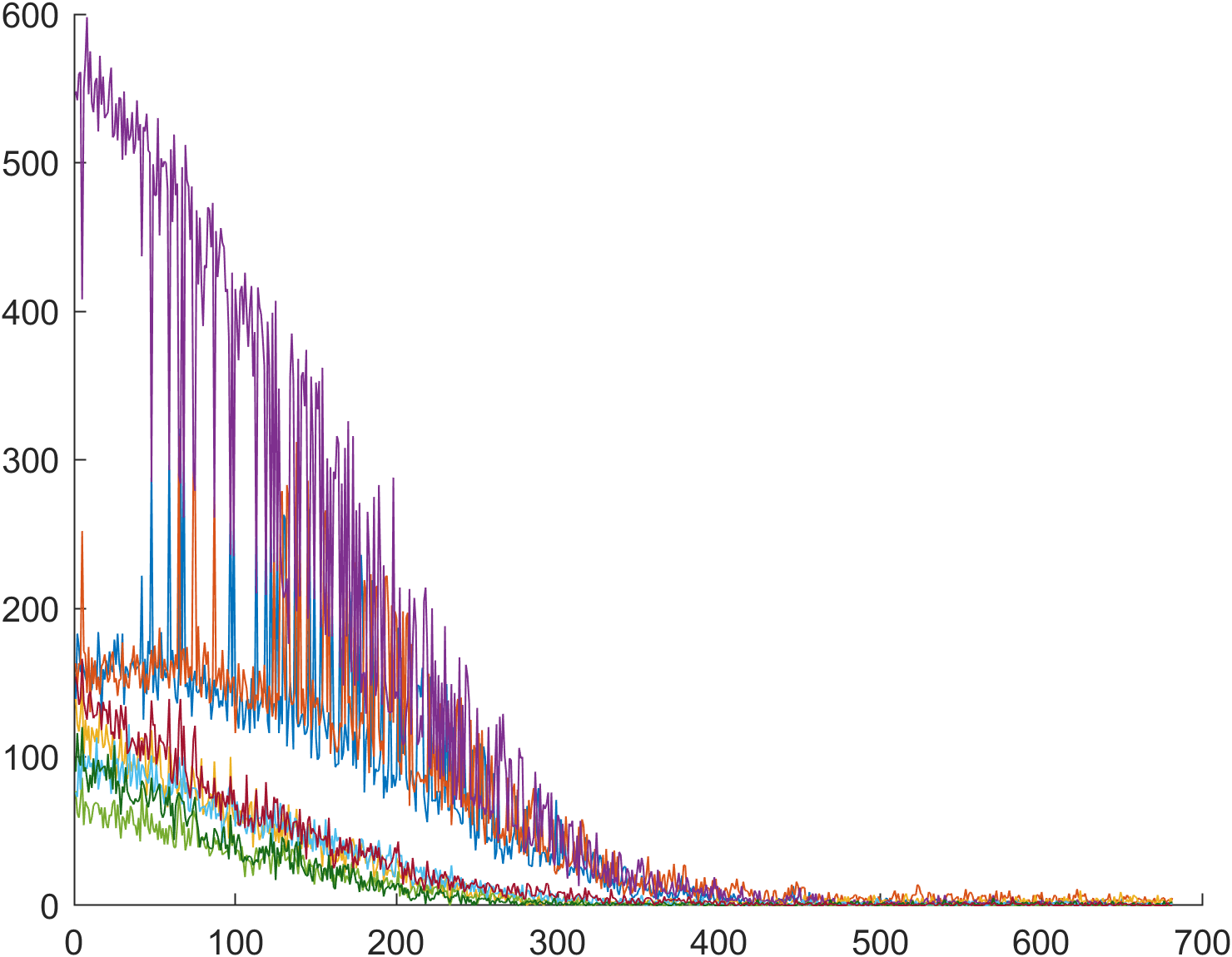}
        \end{subfigure}
    \caption{HHASA$_{\epsilon-G}$}
    \label{fig:E101EG}
    \end{subfigure}
        \hspace{0.2cm}
    \begin{subfigure}{0.2\textwidth}
        \begin{subfigure}{1.10\textwidth}
        \centering
        \includegraphics[width=\linewidth]{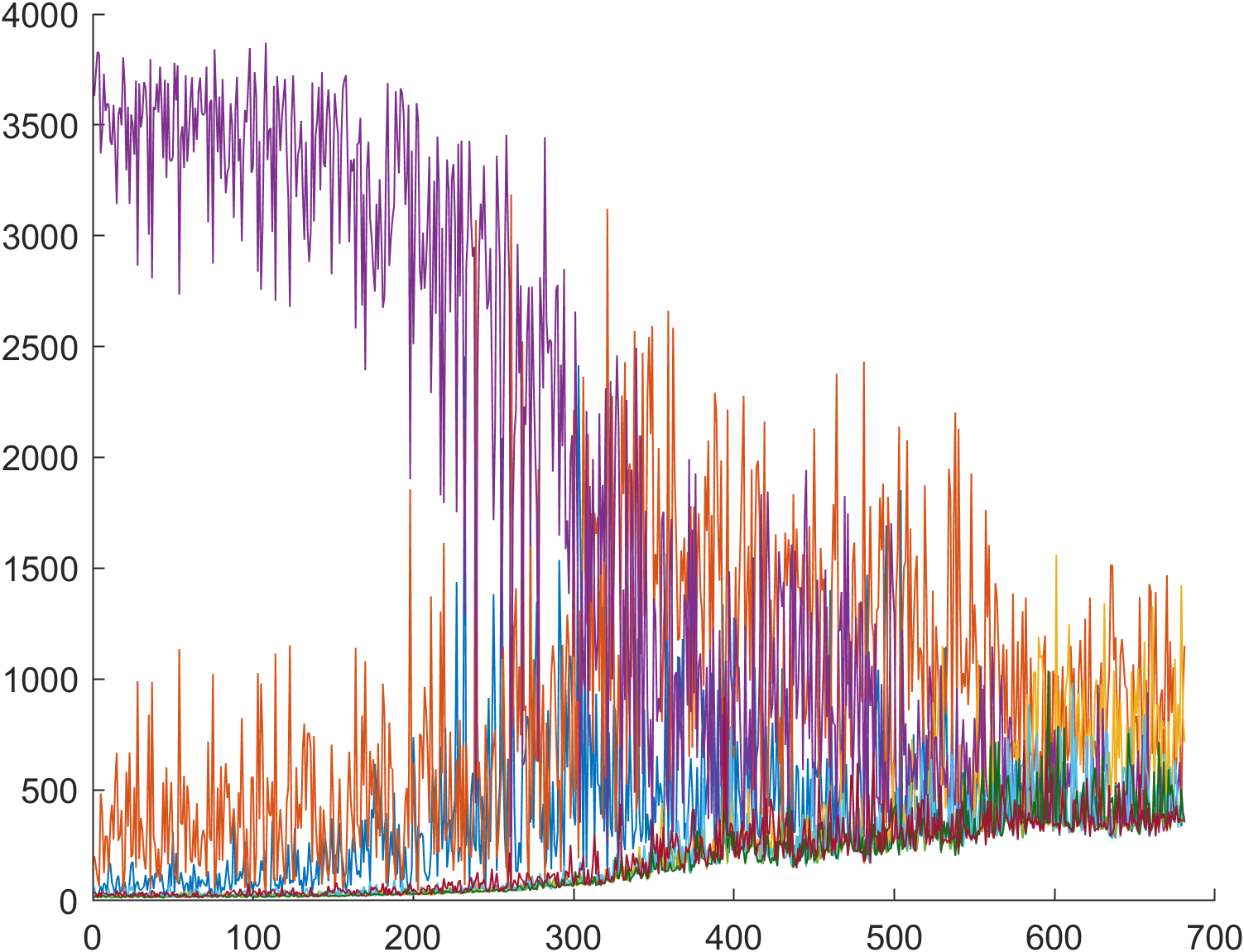}
        \end{subfigure}
        \begin{subfigure}{1.10\textwidth}
        \centering
        \includegraphics[width=\linewidth]{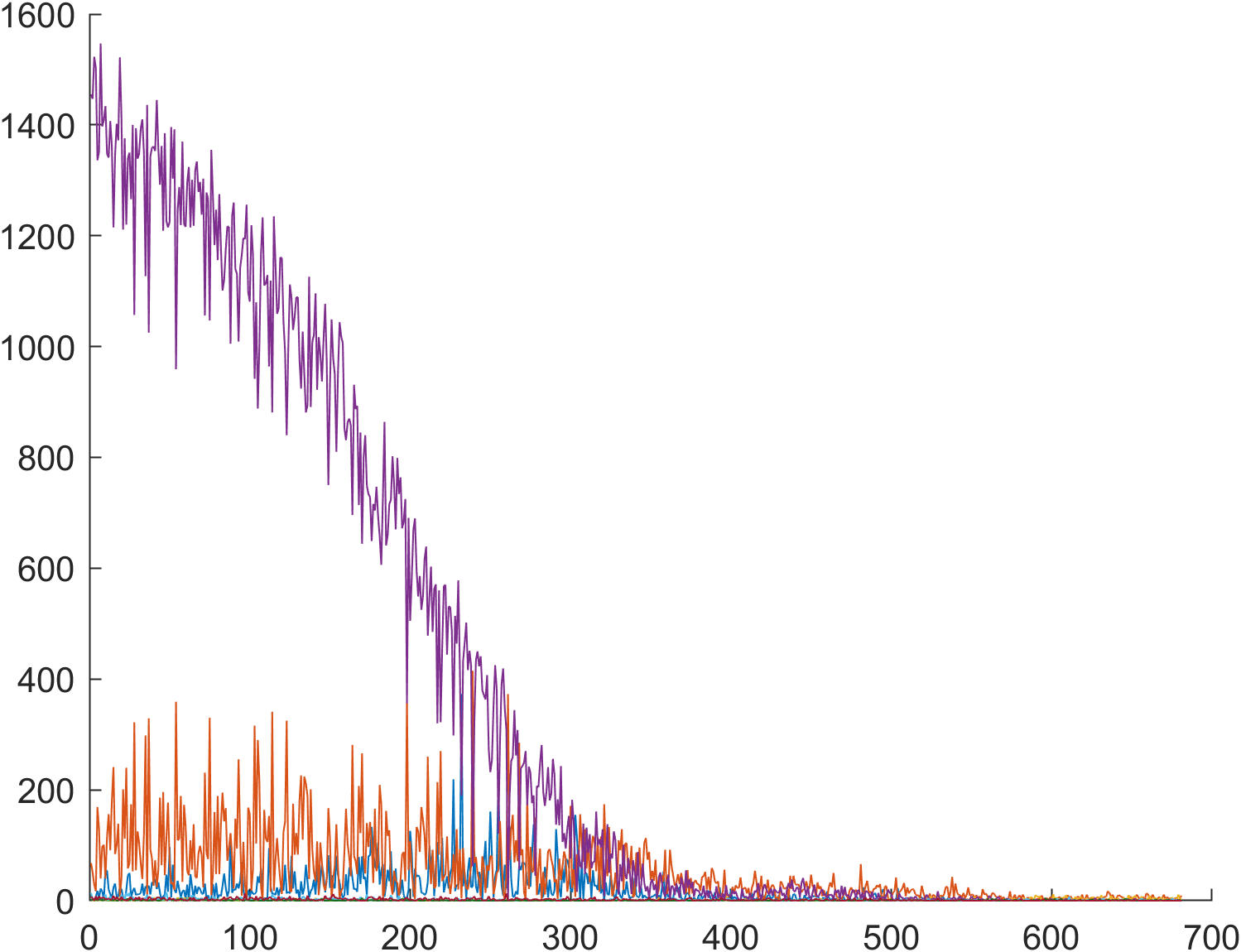}
        \end{subfigure}
    \caption{HHASA$_{TS}$}
    \label{fig:E101TS}
    \end{subfigure}
        \hspace{0.2cm}
    \begin{subfigure}{0.2\textwidth}
        \begin{subfigure}{1.10\textwidth}
        \centering
        \includegraphics[width=\linewidth]{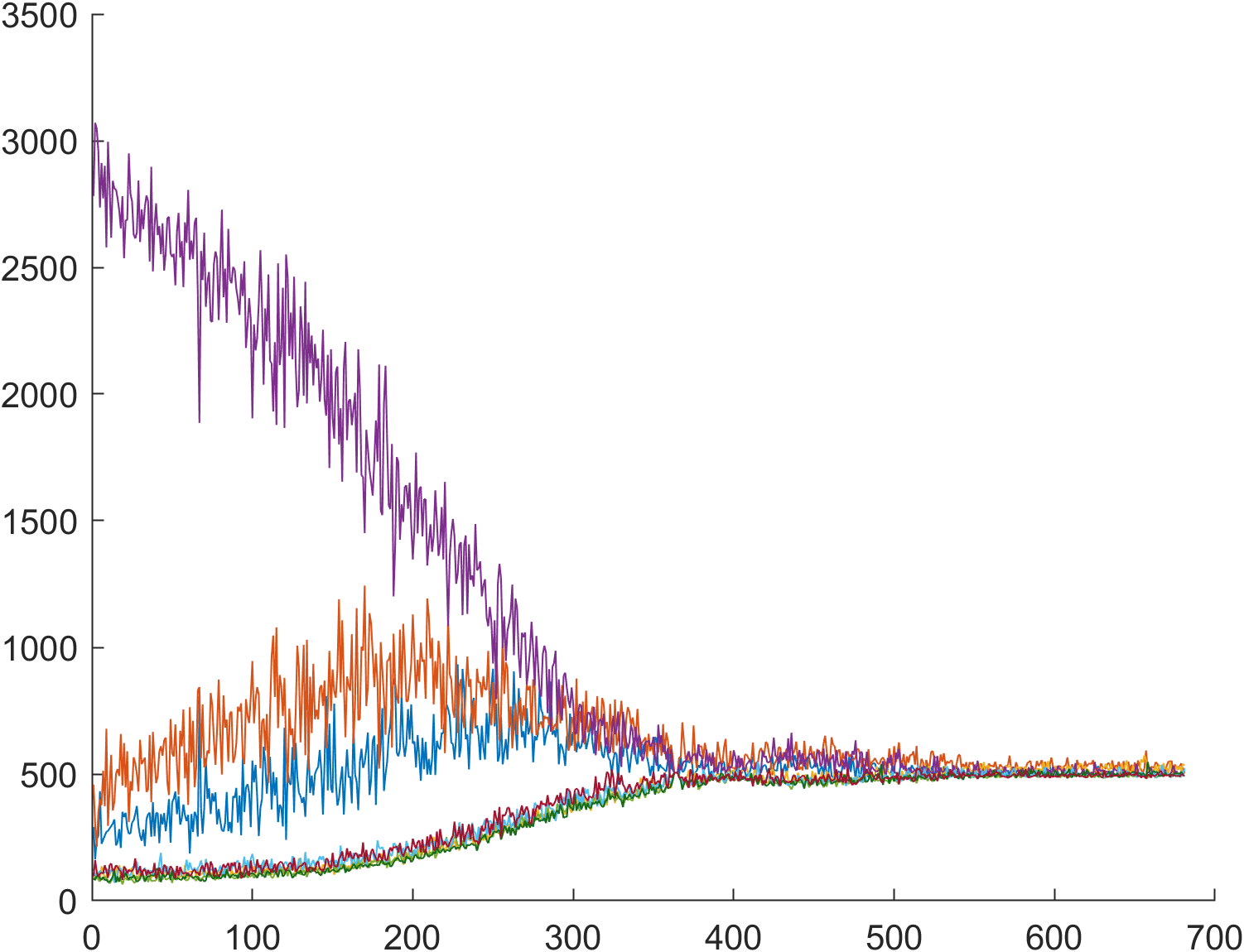}
        \end{subfigure}
        \begin{subfigure}{1.10\textwidth}
        \centering
        \includegraphics[width=\linewidth]{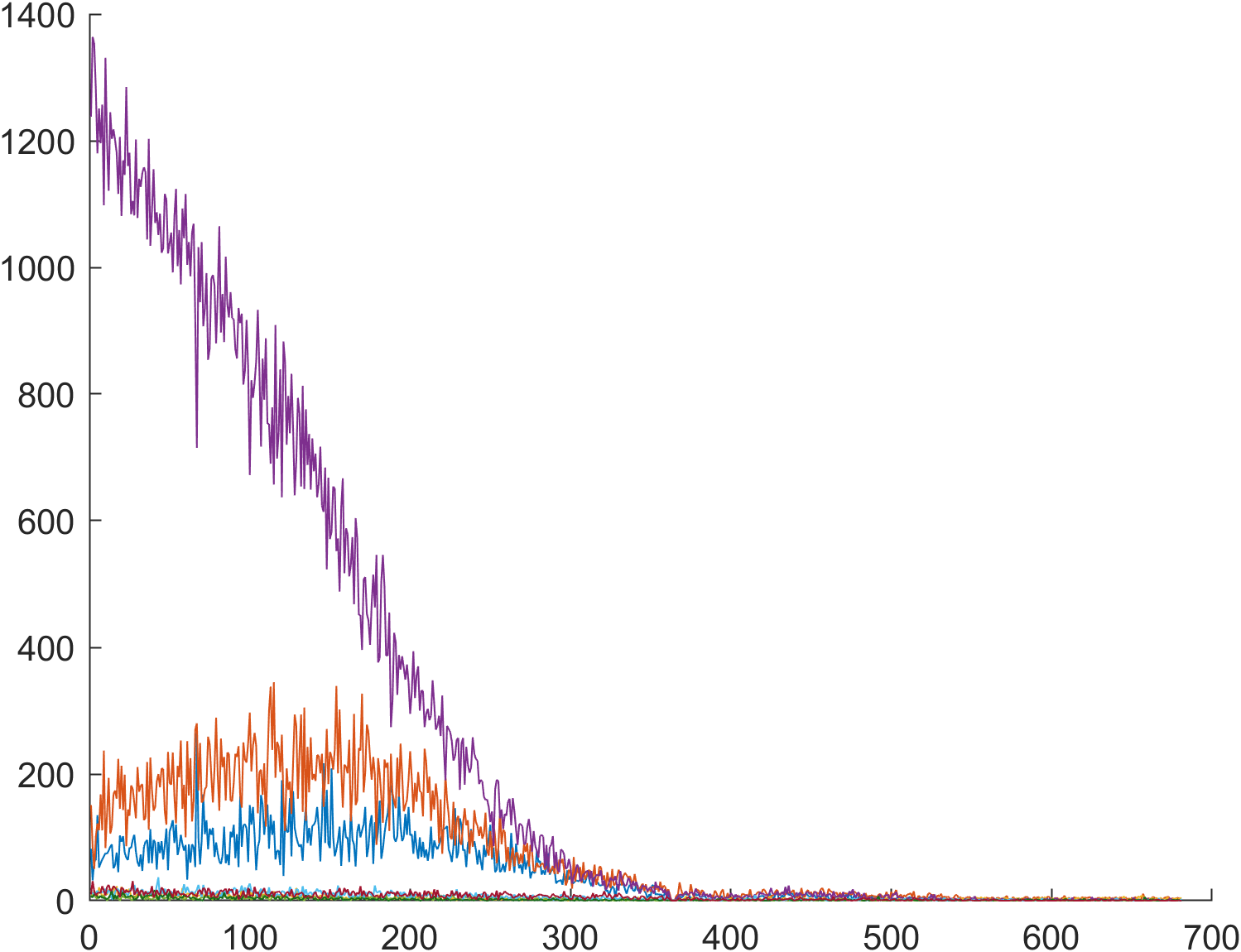}
        \end{subfigure}
    \caption{HHASA$_{UCB1}$}
    \label{fig:E101UCB1}
    \end{subfigure}
\caption{Graphs of the vectors of the selected heuristics and the rewards of all the local searches of the four proposals of this work for the E101 instance.}
\label{fig:E101}
\end{figure}

\begin{figure}[H] 
\centering
    \begin{subfigure}{0.11\textwidth}
        \begin{subfigure}{\textwidth}
        \centering
        \includegraphics[width=\linewidth]{Images/Graficas/Etiquetas2.png}
        \end{subfigure}
    \end{subfigure}
    \begin{subfigure}{0.2\textwidth}
        \begin{subfigure}{1.10\textwidth}
        \centering
        \includegraphics[width=\linewidth]{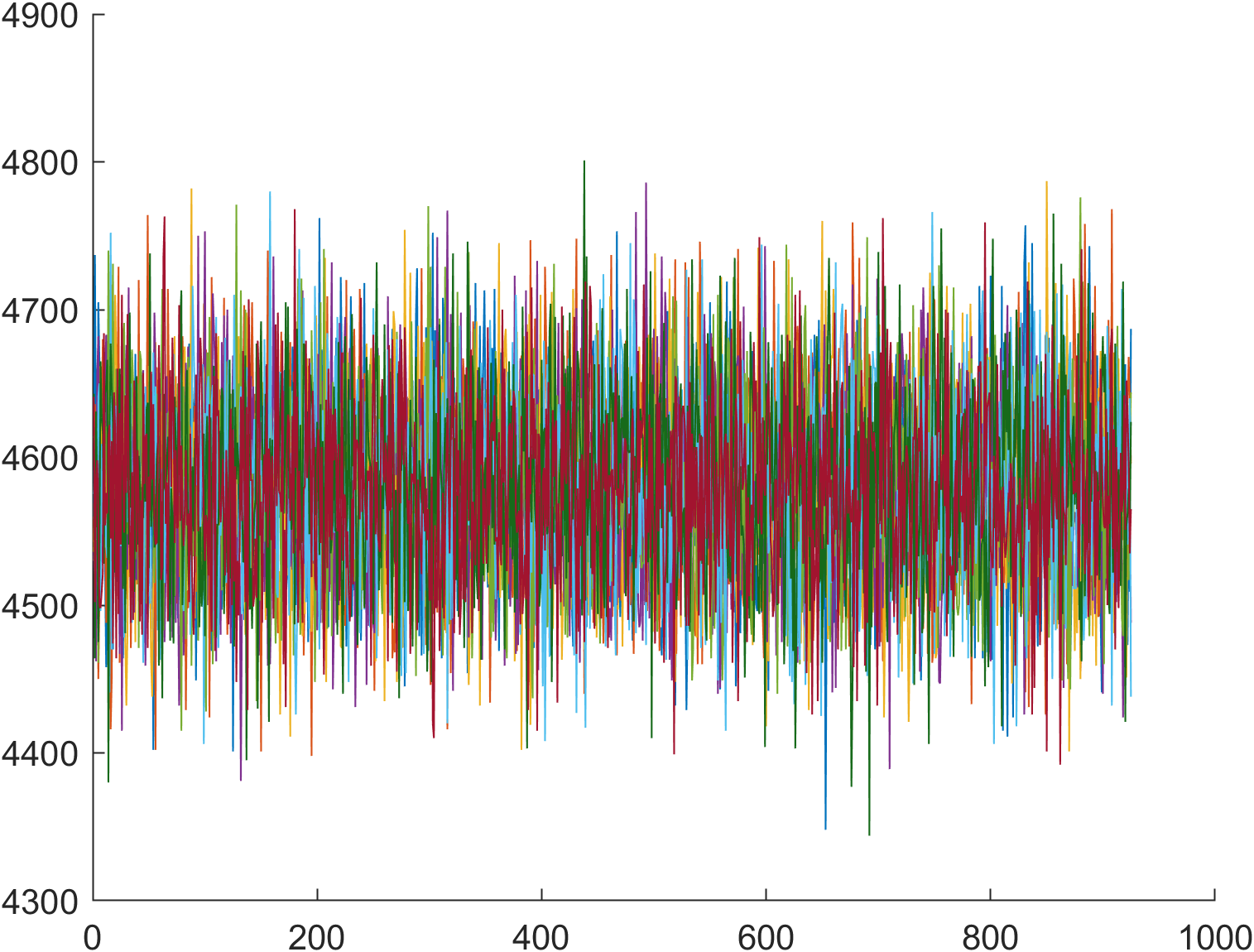}
        \end{subfigure}
        \begin{subfigure}{1.10\textwidth}
        \centering
        \includegraphics[width=\linewidth]{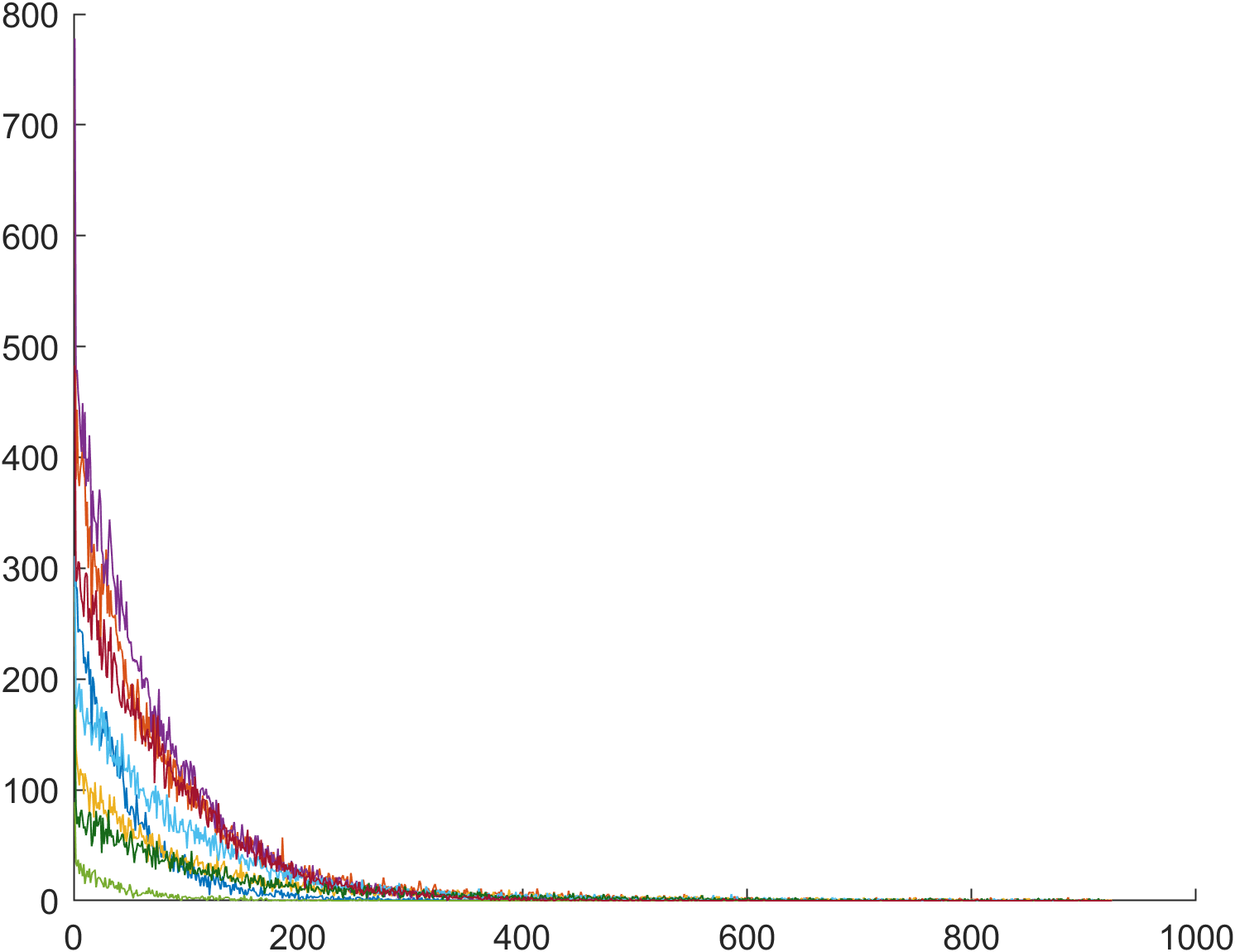}
        \end{subfigure}
    \caption{HHASA}
    \label{fig:X916Rand}
    \end{subfigure}
        \hspace{0.2cm}
    \begin{subfigure}{0.2\textwidth}
        \begin{subfigure}{1.10\textwidth}
        \centering
        \includegraphics[width=\linewidth]{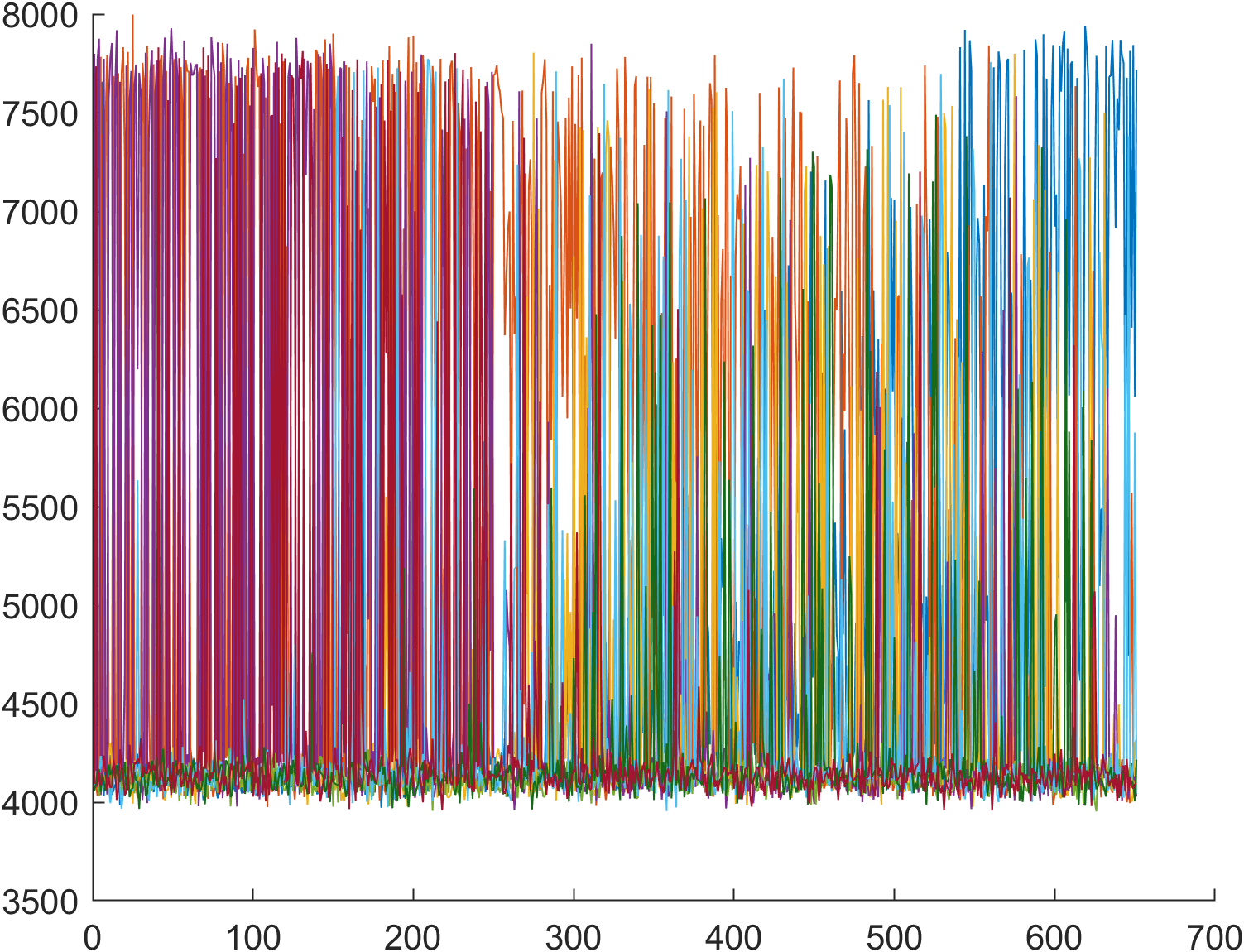}
        \end{subfigure}
        \begin{subfigure}{1.10\textwidth}
        \centering
        \includegraphics[width=\linewidth]{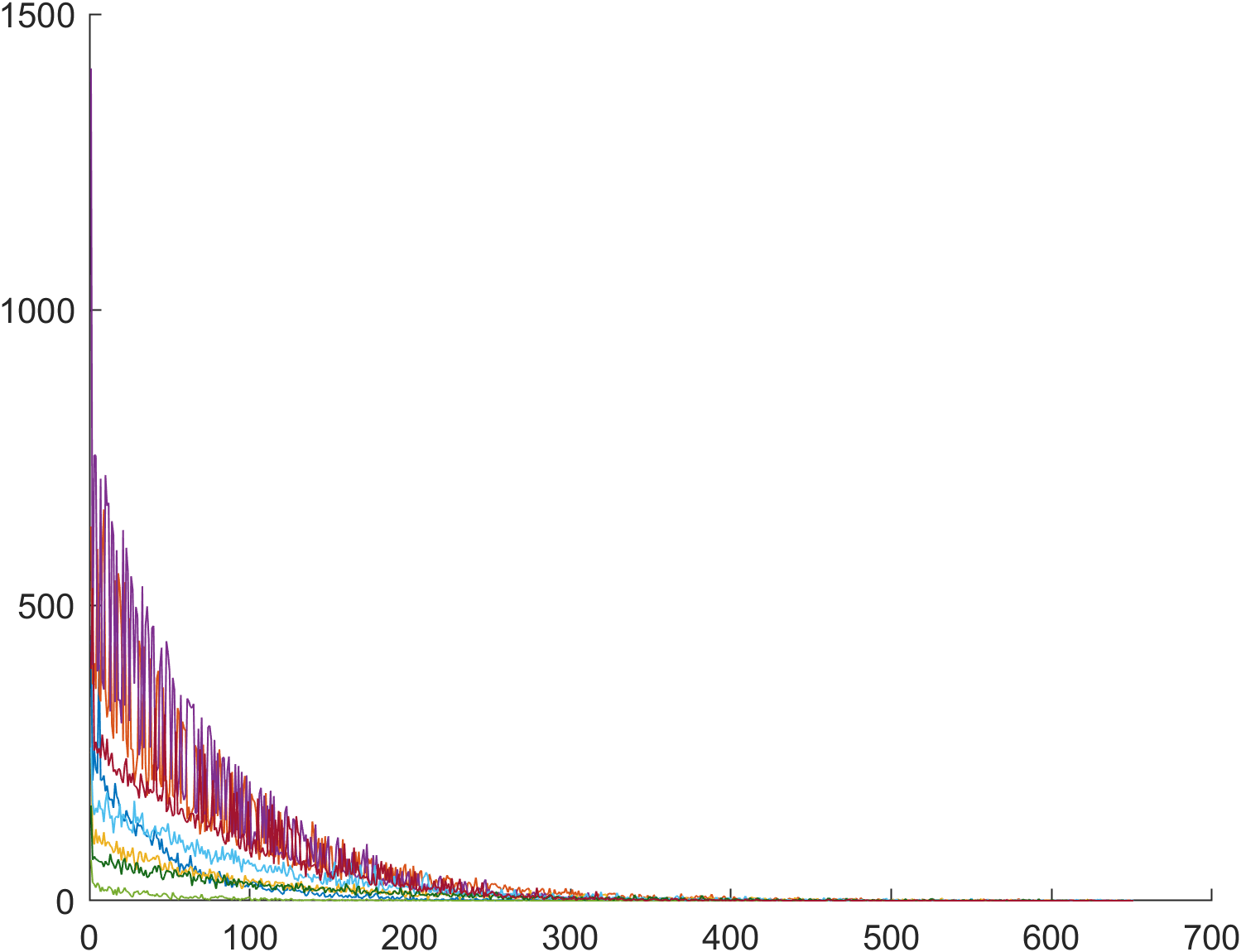}
        \end{subfigure}
    \caption{HHASA$_{\epsilon-G}$}
    \label{fig:X916EG}
    \end{subfigure}        
        \hspace{0.2cm}
    \begin{subfigure}{0.2\textwidth}
        \begin{subfigure}{1.10\textwidth}
        \centering
        \includegraphics[width=\linewidth]{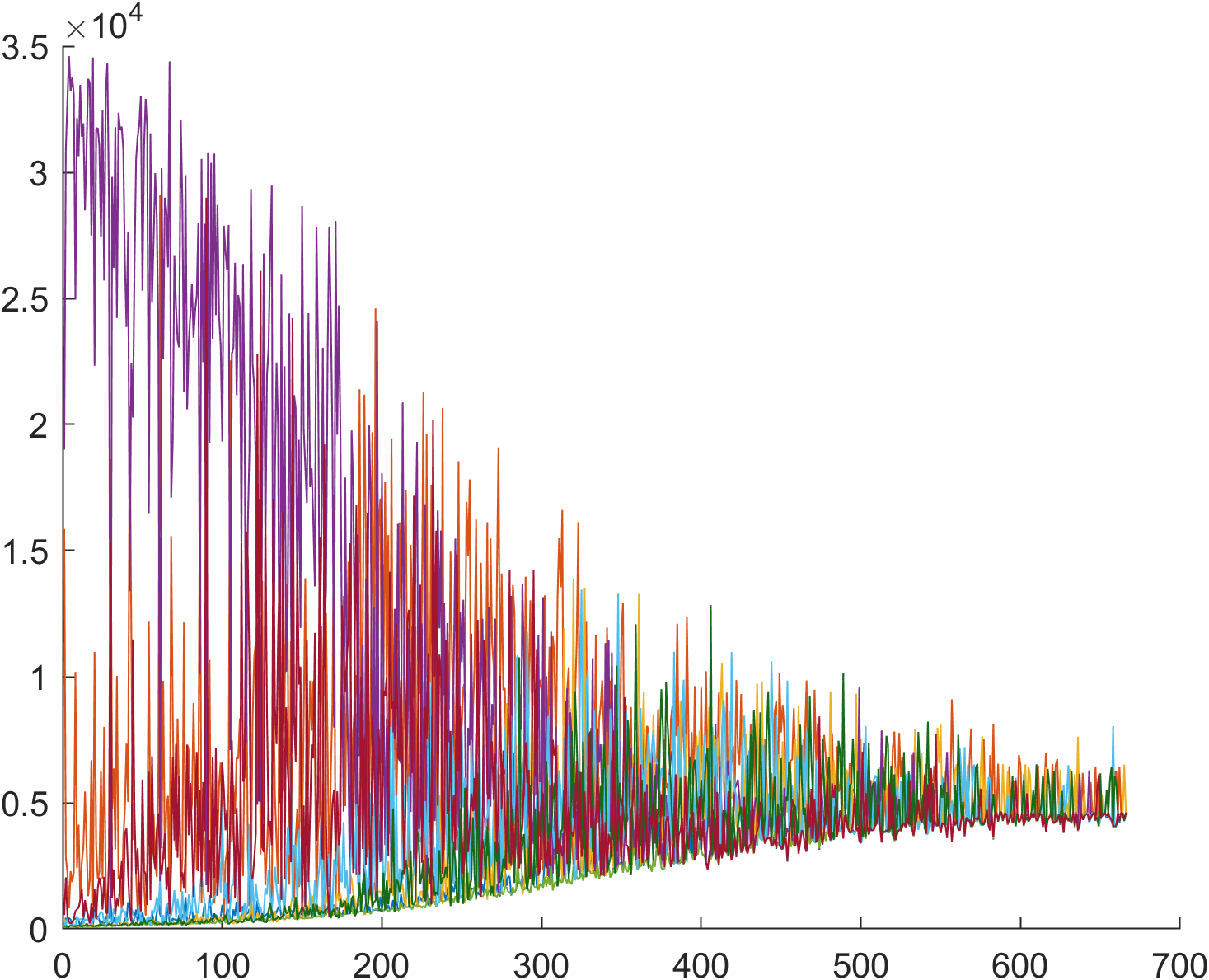}
        \end{subfigure}
        \begin{subfigure}{1.10\textwidth}
        \centering
        \includegraphics[width=\linewidth]{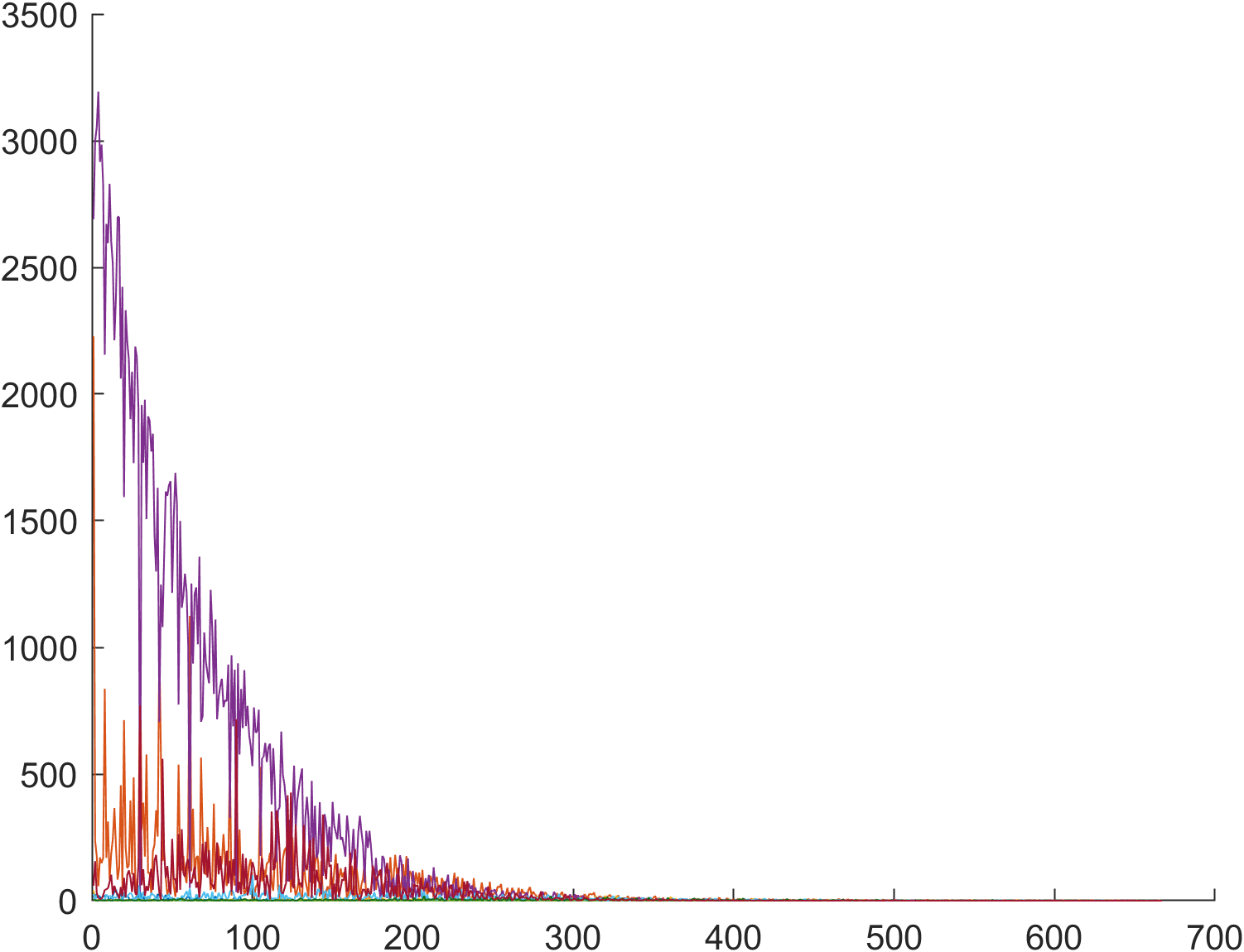}
        \end{subfigure}
    \caption{HHASA$_{TS}$}
    \label{fig:X916TS}
    \end{subfigure}
        \hspace{0.2cm}
    \begin{subfigure}{0.2\textwidth}
        \begin{subfigure}{1.10\textwidth}
        \centering
        \includegraphics[width=\linewidth]{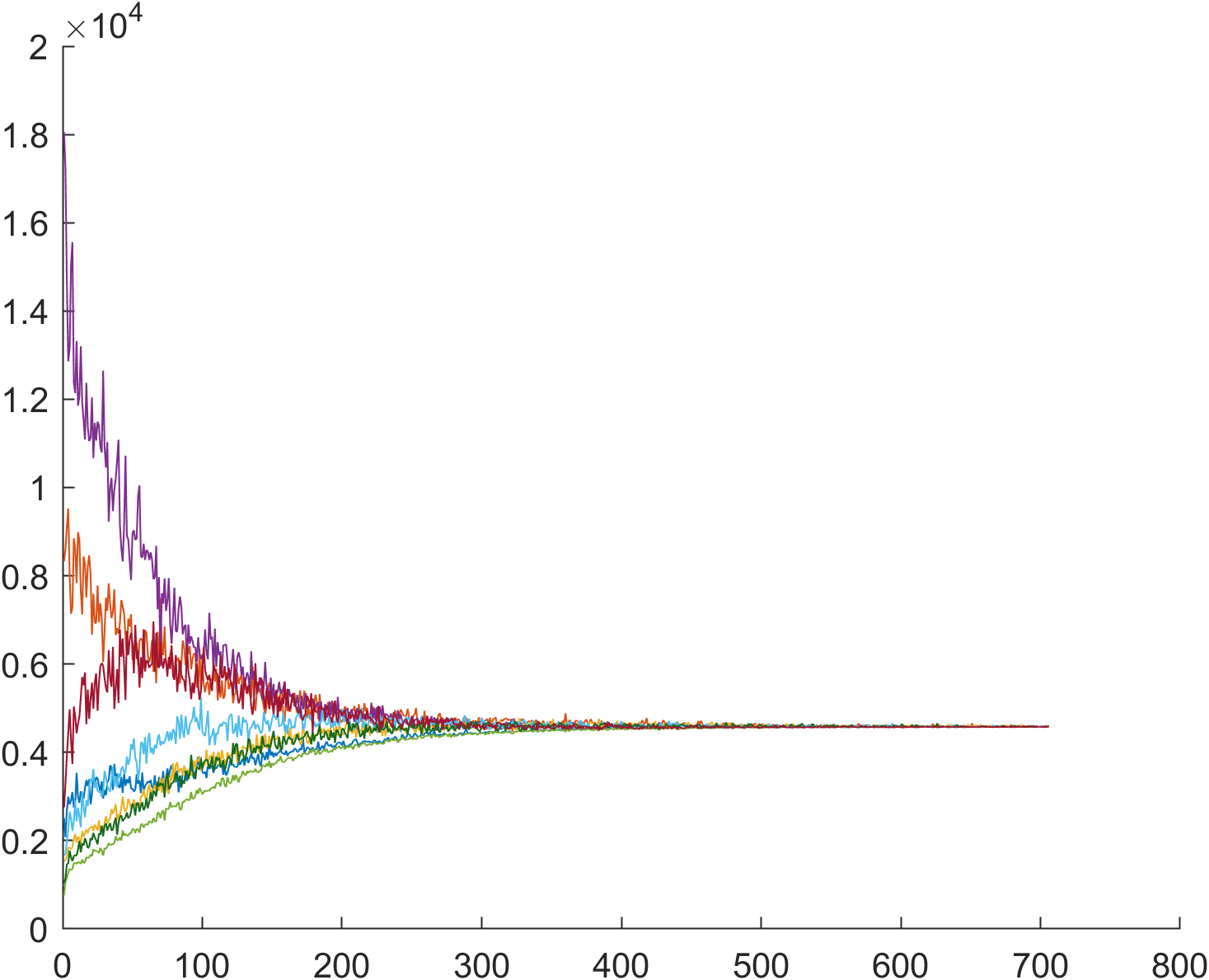}
        \end{subfigure}
        \begin{subfigure}{1.10\textwidth}
        \centering
        \includegraphics[width=\linewidth]{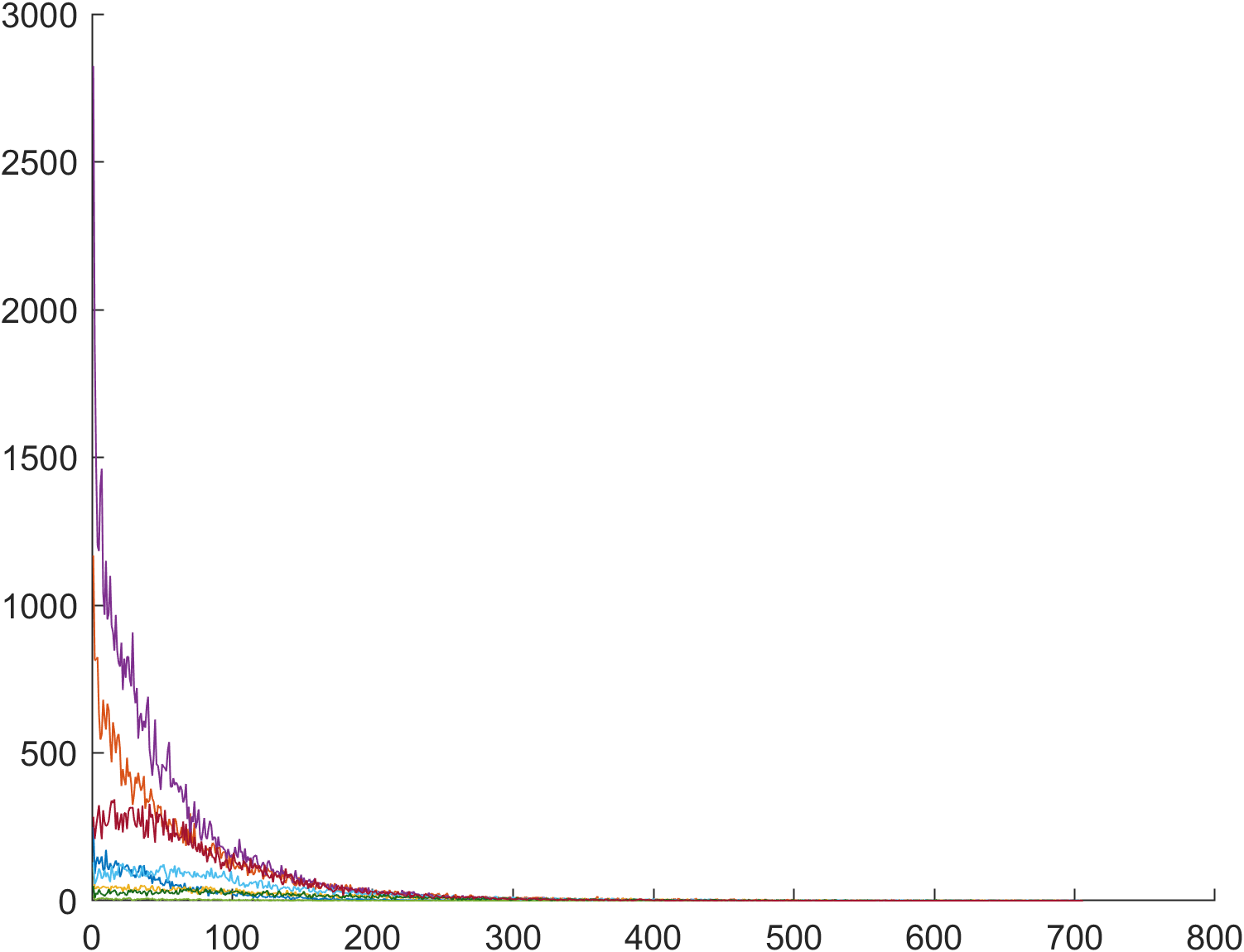}
        \end{subfigure}
    \caption{HHASA$_{UCB1}$}
    \label{fig:X916UCB1}
    \end{subfigure}
\caption{Graphs of the vectors of the selected heuristics and the rewards of all the local searches of the four proposals of this work for the X916 instance.}
\label{fig:X916}
\end{figure}

Figures \ref{fig:E101} and \ref{fig:X916} show the plots of each of the local searches in a run for instances E101 and X916, respectively. The first row of both images indicates the number of heuristics selected for each of the local searches for the four methods. The second row shows the reward vector of the eight heuristics for all local searches.

Figures \ref{fig:E101Rand} and \ref{fig:X916Rand} show the HHASA proposal, and as expected, the heuristic selection vector does not give preference to any of them because it is random. As a result, the reward figures show how well each heuristic performs when they have the same chance of being chosen.

Likewise, Figures \ref{fig:E101EG} and \ref{fig:X916EG} present the vectors of the HHASA$_{\epsilon-G}$ algorithm, which are similar to the proposal that does not use the RL method. The difference existing is more noticeable in the vector of the selected heuristics, where the highest peaks are due to the nature of the $\epsilon$-$G$ algorithm that chooses the heuristic with the highest reward in that local search.

The HHASA$_{TS}$ plots are presented below in Figures \ref{fig:E101TS} and \ref{fig:X916TS}. The Beta distribution used in the Thompson Sampling process shifts from a flat linear to a more realistic probability model of the mean reward as more data is collected. Actions that have been tried infrequently have a higher range of possible values due to their wide dispersion. As a result, a heuristic with a low estimated mean reward that has been tried less times than a heuristic with a high estimated mean may provide a higher sample value, implying that this heuristic probably is selected at that instant of the local search. The $Insertion1$ heuristic is selected more frequently in both cases in the first half of the search process. While in the second half of the search process, the heuristic selection graph begins to behave more like the random mode because the number of times a reward is obtained decreases.

Finally, in Figures \ref{fig:E101UCB1} and \ref{fig:X916UCB1}, the vectors of the proposed HHASA$_{UCB1}$ are presented graphically. The $UCB1$ method has a lower regret level than the Epsilon Greedy and Thompson Sampling methods, so the optimal selection is quickly identified, and the other heuristics are only tested when they have high uncertainty. As it can be observed, in the beginning, it selects the heuristic $Insertion1$ as optimal at the end of the first local searches and gives less number of selections to the other heuristics. However, as the local searches progress, it increases the uncertainty of the heuristic $Insertion1$ and assigns more  probability to the other heuristics to be selected.

\subsection{Comparison of additional energy used}

Table \ref{Tab:Energy} presents the additional energy that the two best HHs proposals and the BACO algorithm have to use. This extra energy is obtained from the difference between the mean values obtained from each algorithm and the smallest distance found for the benchmark instances in Tables \ref{Tab:Small} y \ref{Tab:Large}. This energy consumed is calculated through the multiplication of the difference between the average and the best fitness, which indicates the average and the shortest route lengths, respectively, and it is multiplied by the energy consumption constant $h$, specified in the Table \ref{Tab:Details}. In this table, it can be seen that the difference is not remarkable for small instances. However, as the number of clients increases, the energy difference between the two HHs and the BACO starts to be significantly higher, as it is possible to notice in the X573 instance. Also, in this comparison, the algorithm that presented the lowest total energy difference is the HHASA$_{TS}$.

\begin{table}[H]
\centering
\caption{Difference in energy consumption between the average distance values and the minimum distance found so far.}
\label{Tab:Energy}
\begin{tabular}{c|c|c|c} 
    \hline 
    Instances & HHASA$_{TS}$ & HHASA$_{UCB1}$ & BACO\\ 
    \hline 
    E22 & 0 & 0 & 0\\
    E23 & 0 & 0 & 0\\
    E30 & 0 & 0 & 0\\
    E33 & 0.67 & 0.32 & 2.59\\
    E51 & 8.50 & 6.28 & 0.00\\
    E76 & 2.78 & 3.61 & 0.25\\
    E101 & 8.96	& 9.22 & 12.38\\
    X143 & 313.14 & 330.10 & 130.23\\
    X214 & 116.32 & 170.55 & 129.42\\
    X351 & 272.26 & 282.01 & 114.84\\
    X459 & 289.07 & 284.64 & 164.57\\
    X573 & 340.70 & 328.24 & 3131.15\\
    X685 & 445.30 & 763.15 & 1484.62\\
    X749 & 402.68 & 523.37 & 961.55\\
    X819 & 605.88 & 894.99 & 3641.00\\
    X916 & 924.21 & 1631.86 & 8282.24\\
    X1001 & 461.99 & 662.27 & 1965.04\\ \hline 
    Total &4192.47 & 5890.61 & 20019.89\\
    \hline                  
\end{tabular}
\end{table}

\subsection{Graphical analysis of the solution}

\begin{figure}[H]
\centering
    \begin{subfigure}[b]{0.23\textwidth}
        \includegraphics[width=\linewidth,height=3.5cm]{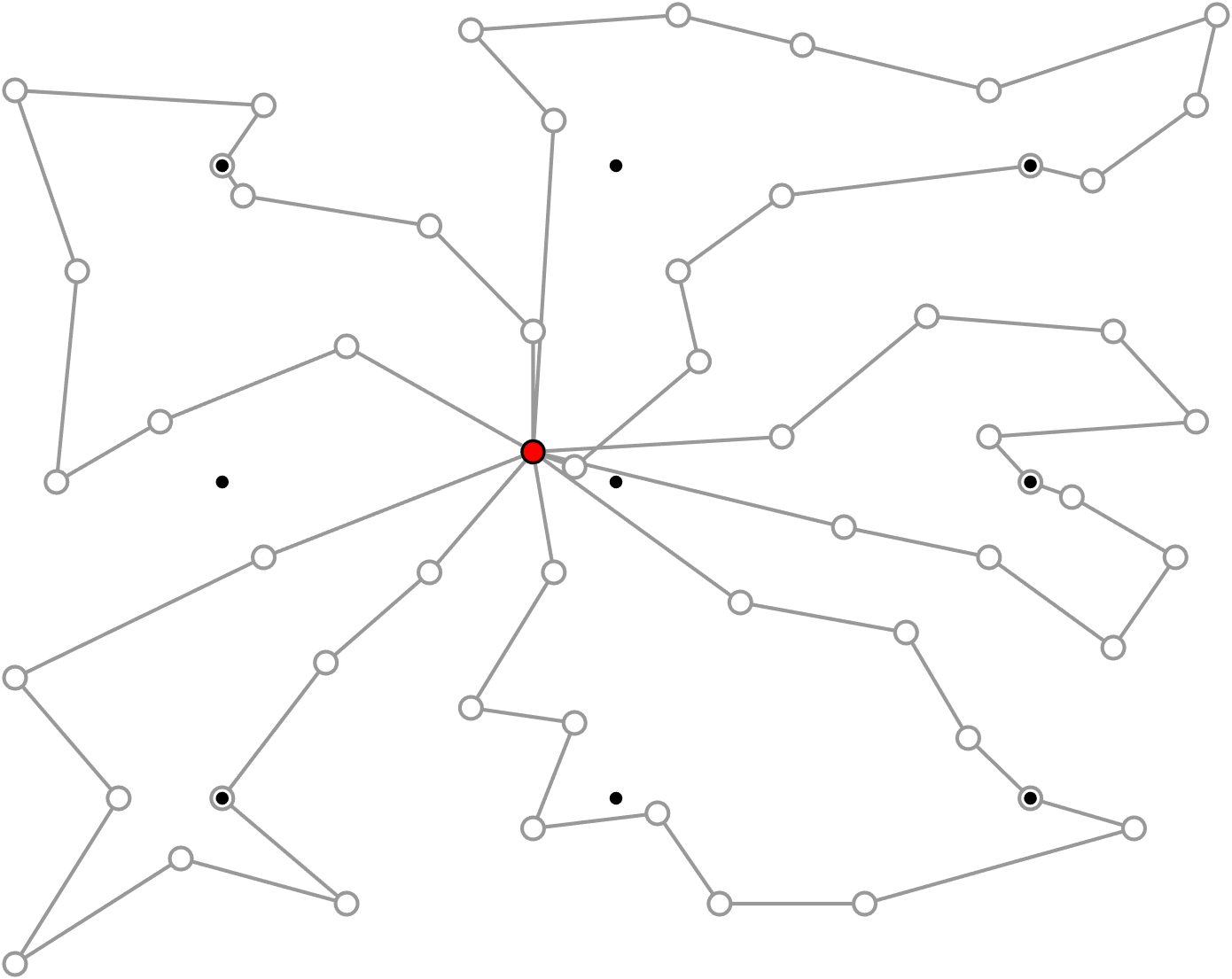}
        \caption{E51}
        \label{fig:DiagramasA}
    \end{subfigure}
    \hspace{0.2cm}
    \begin{subfigure}[b]{0.23\textwidth}
        \includegraphics[width=\linewidth,height=3.5cm]{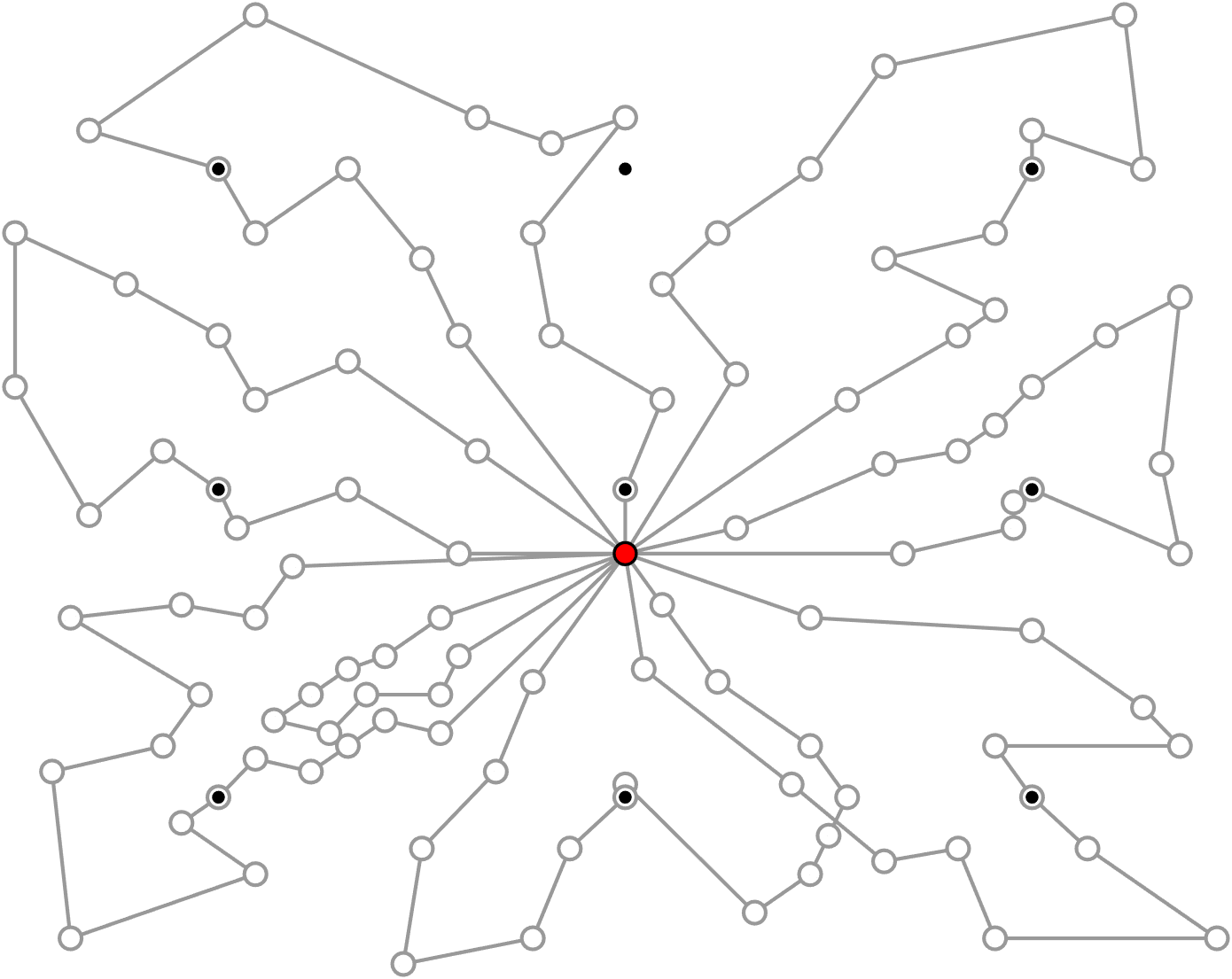}
        \caption{E101}
        \label{fig:DiagramasB}
    \end{subfigure}
    \hspace{0.2cm}
    \begin{subfigure}[b]{0.23\textwidth}
        \includegraphics[width=\linewidth,height=3.5cm]{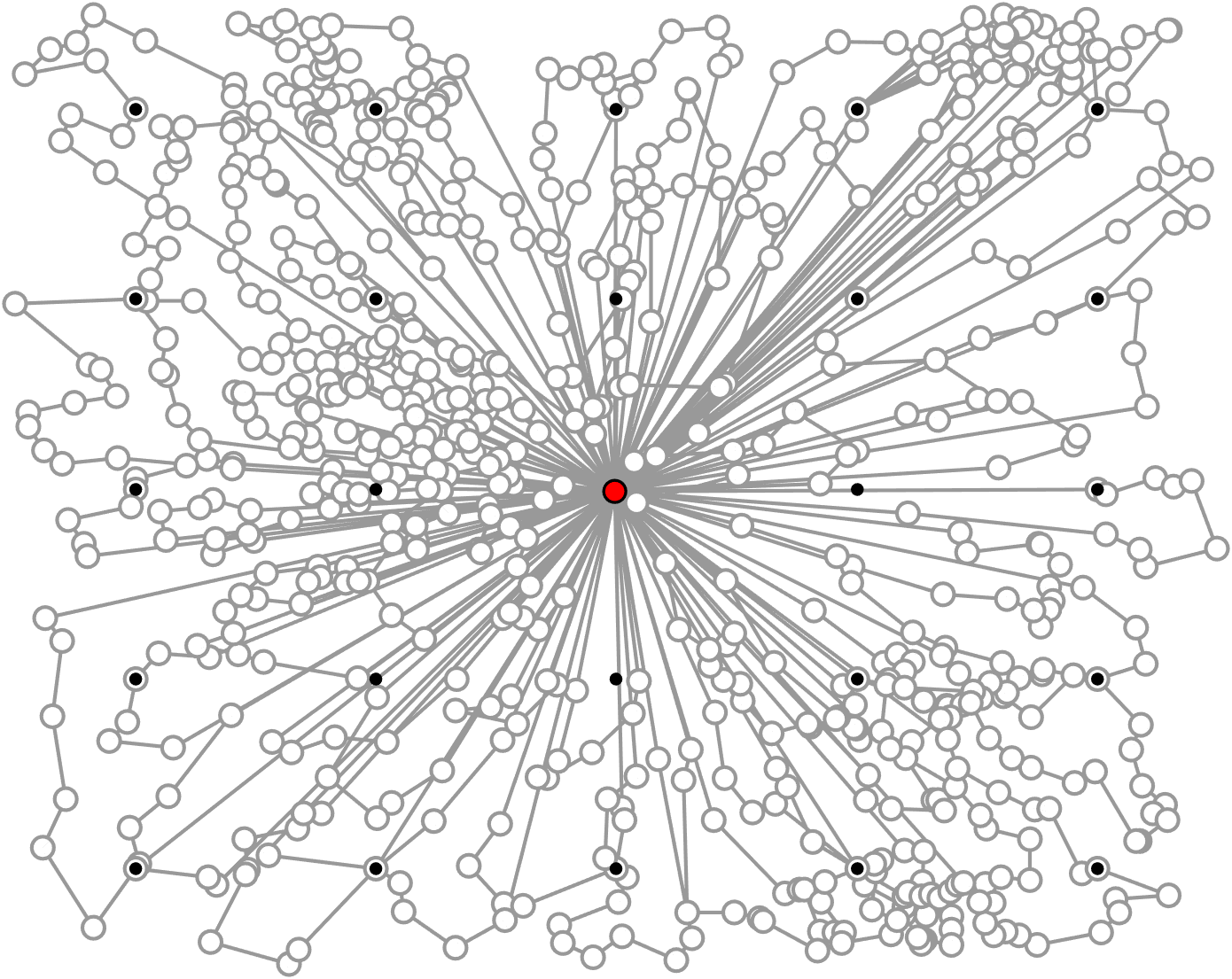}
        \caption{X685}
        \label{fig:DiagramasC}
    \end{subfigure}
    \hspace{0.2cm}
    \begin{subfigure}[b]{0.23\textwidth}
        \includegraphics[width=\linewidth,height=3.5cm]{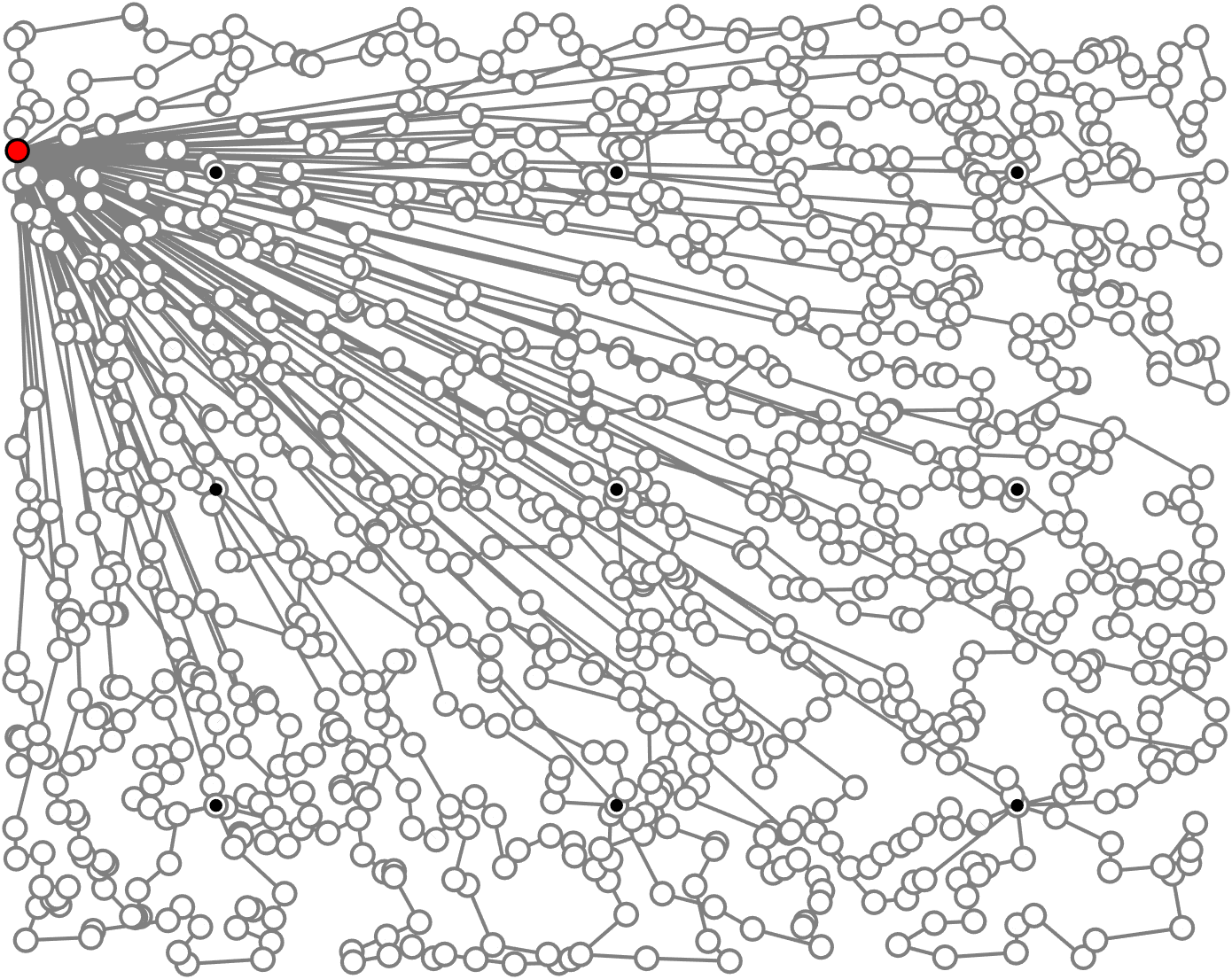}
        \caption{X1001}
        \label{fig:DiagramasD}
    \end{subfigure}
\caption{Solutions generated by HHASA$_{TS}$ for instances E51, E101, X685 and X1001. The symbols $\textcolor[rgb]{1,0,0}{\bullet}$, $\textcolor[rgb]{0.5,0.5,0.5}{\circ}$ and $\bullet$ represent the depot, customers and charging stations respectively.}
\label{fig:Diagramas}
\end{figure}

Figure \ref{fig:Diagramas} shows visually the best solutions generated by applying the HHASA$_{TS}$ algorithm on the CEVRP benchmark in four instances, E51 (\ref{fig:DiagramasA}), E101 (\ref{fig:DiagramasB}), X685 (\ref{fig:DiagramasC}) y X1001 (\ref{fig:DiagramasD}). In Figure \ref{fig:DiagramasA} and \ref{fig:DiagramasB} it is possible to identify the number of routes visually. However, in the routing of large distances such as \ref{fig:DiagramasC} and \ref{fig:DiagramasD}, the routes can not be visualized. Figure \ref{fig:DiagramasB} shows the eight routes the best solution found. In addition, it can be noticed that only one charging station is not used and that one of the routes uses two charging stations in its route. 

On the other hand, it is important to visualize that the proposed HH manages to effectively set up the charging stations to prevent the EV from excessively deviating from the route to the remaining customers. Finally, the figures for instances X685 and X1001 are shown to graphically highlight the difficulty of the benchmark on which the algorithm was tested. 

%% file: 8Conclusions.tex
\section{Conclusions and Future Work}
\label{S:8}

Last-mile logistics has had a significant economic, social, and environmental impact in urban areas as a consequence of the increasing number of vehicles transporting goods. The objective of this work is to develop a methodology for optimizing freight vehicles routes by promoting the use of EVs to increase efficiency and reduce the times and/or costs of last-mile logistics. Therefore, an efficient algorithm called HHASA$_{RL}$ is proposed to find the optimal solutions to high-dimensional CEVRP due to the deficiencies the state-of-the-art methods to deal with this type of instances.

First, a hybridization between the Metropolis criterion of the well-known self-adaptive metaheuristic SA algorithm as a movement acceptance mechanism and the RL algorithm as a heuristic selection mechanism for the design of a HH focused on CEVRP. Second, the experimental results show that the proposed approach applied to the benchmark of the IEEE WCCI2020 competition outperforms all the algorithms found in the state-of-the-art that used the same dataset. In addition, multiple new best-known solutions for high-dimensional instances were found. Finally, it is important to highlight that, the three proposals of HHASA$_{RL}$ algorithms have a more efficient and better performance than the compared algorithms for large instances. However, the algorithm that presents the best performance is the HHASA$_{TS}$ according to the average results and non-parametric tests, which uses the Thompson Sampling method to solve the multi-armed bandit problem.

According to the experiments conducted, although HHASA$_{TS}$ has shown better performance than the state-of-the-art algorithms, there are still multiple lines that deserve more research. As future work, we intend to modify the internal $Adjust Station$ block of the proposed HH with some other novel approach to make the technique of anticipating stops at the charging stations more efficient. In addition, the $RL$ block will be replaced by deep RL methods to make it more robust and test the adaptability and efficiency of using these techniques as a heuristic selection mechanism. 

Finally, it is planned to test this and the new approaches with the above improvements on more complex problems such as the capacitated electric vehicle routing problem with time windows (CEVRPTW), where a fleet of delivery EVs must serve customers with known demand but with opening hours for a single product.